%% file: paper.tex
\documentclass[conference]{IEEEtran}
\IEEEoverridecommandlockouts
\usepackage{cite}

\newcommand{\citet}{\cite}
\usepackage{amsmath,amssymb,amsfonts}
\usepackage{algorithmic}
\usepackage{graphicx}
\usepackage{textcomp}
\usepackage{xcolor}
\usepackage{booktabs}
\usepackage{multirow}
\usepackage{wrapfig}
\usepackage{hyperref}
\usepackage{color}
\usepackage{caption}
\usepackage{subcaption}
\usepackage{url}
\usepackage{enumitem}
\usepackage{xcolor}
\usepackage[capitalise,nameinlink]{cleveref} %nameinlink makes whole word colored
\usepackage[T1]{fontenc}
\usepackage{relsize}

\newcommand{\ourmethod}{TabSketchFM}
\newcommand{\lakebench}{LakeBench}
\newcommand{\spiderlb}{Spider-OpenData}
\definecolor{darkgreen}{rgb}{0,0.5,0}
\setlist[itemize]{align=parleft,left=0pt..1em}

\newcommand{\hk}[1]{{{\color{teal}{[\textbf{HK}: #1]}}}}

\newcommand{\revision}[1]{{{#1}}}
\newcommand{\introparagraph}[1]{\noindent\textbf{#1.}} 

\newcommand\bigexists{%
  \mathop{\lower0.75ex\hbox{\ensuremath{%
    \mathlarger{\mathlarger{\mathlarger{\mathlarger{\exists}}}}}}}%
  \limits}
  
\begin{document}

\title{
TabSketchFM: Sketch-based Tabular Representation Learning 
for Data Discovery over Data Lakes
\thanks{$\star$ Joint first-authors $\blacktriangleright$ Work done while at IBM Research.}
}

\author{
\IEEEauthorblockN{Aamod Khatiwada$\empty^{\star\blacktriangleright}$}
\IEEEauthorblockA{\textit{Northeastern University} \\
khatiwada.a@northeastern.edu}
\and
\IEEEauthorblockN{Harsha Kokel$\empty^\star$}
\IEEEauthorblockA{\textit{IBM Research} \\
harsha.kokel@ibm.com}
\and
\IEEEauthorblockN{Ibrahim Abdelaziz}
\IEEEauthorblockA{\textit{IBM Research} \\
ibrahim.abdelaziz1@ibm.com}
\and
\IEEEauthorblockN{Subhajit Chaudhury}
\IEEEauthorblockA{\textit{IBM Research} \\
subhajit@ibm.com}
\and
\IEEEauthorblockN{Julian Dolby}
\IEEEauthorblockA{\textit{IBM Research} \\
dolby@us.ibm.com}
\and
\IEEEauthorblockN{Oktie Hassanzadeh}
\IEEEauthorblockA{\textit{IBM Research} \\
hassanzadeh@us.ibm.com}
\and
\IEEEauthorblockN{Zhenhan Huang$\empty^\blacktriangleright$}
\IEEEauthorblockA{\textit{Rensselaer Polytechnic} \\ \textit{Institute}\\ 
huangz12@rpi.edu}
\and
\IEEEauthorblockN{Tejaswini Pedapati}
\IEEEauthorblockA{\textit{IBM Research} \\
tejaswinip@us.ibm.com}
\and
\IEEEauthorblockN{Horst Samulowitz}
\IEEEauthorblockA{\textit{IBM Research} \\
samulowitz@us.ibm.com}
\and
\IEEEauthorblockN{Kavitha Srinivas}
\IEEEauthorblockA{\textit{IBM Research} \\
kavitha.srinivas@ibm.com}
}


\maketitle
\input{sections/abstract}
\input{sections/introduction2}
\input{sections/related_work}

\revision{
\input{sections/tabsketchfm2}
}
\input{sections/pretraining}
\input{sections/finetuning}

% \input{sections/benchmark_description}
% hk: Commenting out lakebench 

\input{sections/experiments}
\input{sections/conclusion}

% \bibliographystyle{ACM-Reference-Format}
% \bibliography{references}
\bibliographystyle{IEEEtran}
\bibliography{paper}

\end{document}

%% file: sections/abstract.tex
\begin{abstract}
Enterprises have a growing need to 
identify
relevant tables in data lakes; e.g. 
tables that are unionable, joinable, or subsets of each other.   
Tabular neural models can be helpful for 
such 
data discovery
tasks.
%, 
%but their use in data discovery has not been explored.
%aamod: I think we should not make this claim. We are being criticized that there are neural models for table discovery. We have enough novelty.
In this paper, we present \ourmethod{}, a neural tabular model for data discovery over data lakes. First, we propose novel pre-training: a sketch-based approach 
to enhance the effectiveness of data discovery in neural tabular models. Second, we finetune the pretrained model 
%on a suite that we previously developed called \emph{\lakebench{}} to help build fine tuned models
for identifying unionable, joinable, and subset table pairs and show significant improvement over previous tabular neural models. Third, we present a detailed ablation study to highlight which sketches are crucial for which tasks. Fourth, we use these finetuned models to perform table search; i.e., given a query table, find other tables in a corpus that are unionable, joinable, or that are subsets of the query.  Our results demonstrate significant improvements in F1 scores for search compared to state-of-the-art techniques.  Finally, we show significant transfer across datasets and tasks establishing that our model can generalize across different tasks and over different data lakes. 
% \hk{talk abt ablation?}
\end{abstract}

%% file: sections/introduction2.tex
\section{Introduction}
\label{section:introduction}
Enterprises store critical data in {\em data lakes}, 
large repositories of tabular data, for both governance and analytics~\cite{2019_nargesian_data_lake_management}. It is essential to effectively find relevant tables (e.g., joinable, unionable, subsets) within lakes for reporting, decision-making,  statistical analysis, and more~\cite{Khatiwada2022_integrating_datalake, 2023_khatiwada_dialite}.
For instance, 
\emph{Unionable table search} helps to augment an existing table with new rows, and to enrich analytics over it~\cite{Nargesian2018_Table_union}.
\emph{Joinable table} search is useful to run analytics that needs data in columns from multiple tables, e.g., to identify how many customers in some region purchased a product due to an email campaign. 
Also, identifying \emph{subsets} or \emph{supersets} of a table is essential, for example, when searching for potential copies of personal data to comply with the General Data Privacy Regulation
of the European Union\footnote{\url{https://gdpr-info.eu}}. 
Motivated by such use cases, we focus on the problem of identifying tables from the data lakes that can be \emph{unionable, joinable, and/or subsets of each other}
~\cite{Nargesian2018_Table_union, 2019_zhu_josie, Khatiwada2022_SANTOS}.
Specifically, in this work, we propose a pre-trained tabular model that could help in enhancing the effectiveness of data discovery 
% tasks. 
techniques.

Central to the problem of identifying relevant tables in the data lakes is identifying similar columns in tables.
%relevant tables from data lakes.
%is the issue of finding similar columns or tables.  
However, the columns in a table could be ambiguous. For instance, a column called \texttt{Age} is different when it is in a table about buildings from when it is in a table about people. Furthermore, values within a column \texttt{Age} play a key role in determining semantics; \texttt{Age} of buildings in New York are unlikely to match \texttt{Age} of historical buildings in Egypt. Contextualizing columns could be useful in resolving such ambiguities.
Language Models (LMs) are shown to be successful in contextualizing words in a sentence ~\cite{2019_devlin_bert}.
Consequently, we use them to contextualize columns and use such contextualization to determine similar columns or tables.

Existing pre-trained tabular models~\cite{2020_deng_turl, Iida2021_TABBIE, Wang2021_TUTA, Yin2020_TABERT} based on LMs~\cite{2019_devlin_bert} do not focus on identifying relevant tables. Instead, they focus on language-style tasks such as fact checking~\cite{Wang2021_TUTA}, question answering~\cite{Herzig2020_TAPAS}, converting natural language to SQL~\cite{Yin2020_TABERT},
%table retrieval for a given %question~\cite{Herzig2020_TAPAS, Yin2020_TABERT}, 
table cell content population~\cite{2020_deng_turl}, and table metadata prediction~\cite{Iida2021_TABBIE, 2020_deng_turl}. In addition, input to these models is the actual cell values or a row because of the focus on language-style tasks. However, this has two limitations for understanding relevant tables: (a) only a small number of values can be passed in as input because of severe limits on token length in encoder-only models that can generate embeddings for search (e.g. 512 tokens in BERT~\cite{2019_devlin_bert}), and this potentially limits the information available to the model to understand the input table
{\footnote{
\label{footnote:decoder_only}
\revision{Decoder only models have larger context lengths.  However, it has been difficult until recently to use decoder-only models for embeddings \cite{behnamghader2024llmvec}.  Moreover, these models are expensive to deploy and provide only a minimal improvement in accuracy over encoder-only models such as BERT.  This is why embeddings for search are mostly based on BERT-style models.  Further, even with large context lengths, it is not practical to serialize tables with millions of rows as sentences.}}
; (b) treating numerical values in the table as text tends to lose their semantics; it may be better to pass numerical information directly as vectors into the model.

To address these limitations on input length and numerical representation, 
we introduce a novel transformer-based tabular model---\emph{\ourmethod{}}---exploiting both table metadata and cell values that creates different sketches over them instead of linearizing cell values. Specifically, we abstract column cell values using data sketches, as has been used in the data discovery literature (e.g., MinHash sketches ~\cite{2016_zhu_lsh_ensemble}, and numerical sketches ~\cite{2021_santos_correlation_sketches}).  Such sketches capture tabular features, and we input them into the neural models. Inputting numerical sketches bypasses feeding table cell values to the model, addressing the token limit issue, as well as the numerical representation of cell values.

\revision{
After pretraining using sketches, we finetune the models for individual data discovery tasks such as union, join, and finding subsets. The embeddings given by the finetuned models are then used for the respective retrieval tasks.
While some in the literature \cite{2023_fan_starmie} have used self supervision for fine tuning, self supervision applies better for some tasks such as unions compared to say joinability.  To finetune models for the data discovery tasks, we used a suite of 8 benchmarks across the 3 tasks \emph{\lakebench} that we have open sourced for use \cite{lakebench}.  On the fine tuning tasks, \emph{\ourmethod{}} outperformed other neural methods, by as much as 55\% on F1.  We then used the fine tuned models for search tasks, and showed that \emph{\ourmethod{}} outperforms the state-of-the-art neural and traditional methods by as much as 70\% on F1. }

Importantly, we also compare the performance of specialized traditional and neural models for data discovery against an off-the-shelf pretrained model that encodes column values as a single sentence.  To our surprise, the off-the-shelf model performs much better than existing systems - in fact, for certain tasks such as union, it is sufficient to use that model for discovering unionable tables; our tabular specific model does worse, as do all other models that target unionability.  The insight from this finding is that while some data discovery tasks need values to match across columns (e.g. join or subsets), for union, it is sufficient when columns have the same semantic meaning; indeed there is no need for any value overlap.  Sentence embeddings capture those semantics well. Because certain tasks seem to benefit from these sentence embeddings (e.g. union), and other do not (e.g., subset), we propose a way to concatenate the embeddings from sentence encoders with tabular model embeddings, and show that one can augment tabular embeddings with value embeddings to improve performance, at least in some cases.

In summary, our contributions are as follows:
\begin{itemize}
\item \textbf{A novel tabular pretrained model.} We are the first to present a sketch-based tabular pretrained model, \emph{\ourmethod{}}, which contextualizes columns based on their data sketches and tabular metadata. The sketches help to better handle input token limits and numerical cell values. To achieve this, we modified the text-based transformer architecture to take in numerical and MinHash sketches as input.  A purely sketch based approach though misses the opportunity to include table semantics from column values.  Given that the inputs to the transformer architecture in our system are numeric vectors, it is difficult to combine these numeric inputs with tokens derived from cell values.  We therefore combine the sketch based model embeddings with 
% off the shelf embeddings of 
column value embeddings from existing pretrained sentence transformer models.  We show that this combination boosts performance on some  data discovery tasks.

\iffalse
\hk{reword this contribution or drop it
\item \textbf{Finetuning benchmarks for identifying relevant tables.}
We had developed \emph{\lakebench} earlier, a collection of 8 benchmarks that can be used to build cross encoder models to identify relevant tables. We show how one might use such benchmarks to build fine tuned models for data discovery. }
\fi

\item \textbf{Finetuning models for downstream tasks.} We develop cross-encoders from pretrained \ourmethod{}, which we fine-tune for a range of downstream tasks from the LakeBench collection~\cite{lakebench}.  We fine tune not only \emph{\ourmethod{}} but other tabular pretrained models to see if they can perform well on these downstream tasks and present the comparision.

% \fi 

%\item \textbf{Benchmarks for data discovery learning.} We develop \emph{\lakebench}, 8 benchmarks for training models for data discovery style tasks.  The tables are drawn from enterprise and open data lakes, and tables generated using knowledge graphs. Each benchmark comes with train, test, and validation sets covering a variety of tasks such as binary classification, multi-class classification, and regression. The benchmarks cover unionability, joinability and subsets.

%\item \textbf{A novel tabular model.} We present \emph{\ourmethod{}}, a novel tabular pretrained model that is based on sketches and tabular metadata to contextualize columns.  The use of sketches is designed to better handle input token limits, and numerical cell values. To do so, we modified the text based transformer architecture to include inputs from numerical sketches and MinHash sketches. We show that \emph{\ourmethod{}} outperforms existing neural models for determining table or column similarity, which suggests that abstraction of cell values and proper treatment of numerical values is important.

\item \textbf{Detailed ablation study.} %Detailed ablation studies 
We assess the effectiveness of different %types of 
sketches for identifying different types of table relevance. Our study presents interesting insights such as Minhash sketches of column values are crucial for identifying joins, but numerical sketches are crucial for identifying subsets; once again highlighting the value of using sketches for table representation.

\item \textbf{Demonstration of \ourmethod{} on search.}  We compared the performance of \emph{\ourmethod{}}'s fine tuned models against neural and non-neural baselines. In the case of join search, \emph{\ourmethod{}} outperformed most other systems geared for join search in terms of F1 scores.
% , but tellingly, the sentence encoding model also achieved an F1 score of 92\%.
% One other system JOSIE achieved performance comparable to ours for joins but this system creates an inverted text index on exact cell values, so the approach is much less likely to make errors compared to \ourmethod{} which represents the column as a single compressed vector.
% }
% In addition, the model's performance was robust against transfer across datasets and tasks.  
\item \textbf{Illustrate generalization.} Finally, we illustrate the main advantage of learning a pretrained foundational model by showing the generalization of search performance across tasks. For example, we fine tune the pretrained model on join identification on one dataset and use it for union search on another dataset. Our evaluation shows that our finetuned models generalize well.
\end{itemize}

% \hk{Transfer as contribution???}

%All datasets and code to create the models will be open-sourced upon the acceptance of the paper.

%% file: sections/related_work.tex
\section{Related Work}
\label{section:related_work}

\introparagraph{Tabular Pretrained Models} 
Neural models are pretrained on tabular datasets to either recover masked tokens in the table or to detect corrupt values~\cite{2020_deng_turl, Iida2021_TABBIE, Wang2021_TUTA, Yin2020_TABERT}. \citet{10.1162/tacl_a_00544} surveys 
neural 
models for tabular data representation and their applications.
\citet{2020_deng_turl} combined the Masked Language Model (MLM) objective from BERT~\cite{2019_devlin_bert} with a novel Masked Entity Recovery to train TURL. \citet{Wang2021_TUTA} use MLM with novel Cell-level Cloze and Table Context Retrival for TUTA. \citet{Yin2020_TABERT} used Masked Column Prediction and Cell Value Recovery for TABERT while \citet{Iida2021_TABBIE} repurposed ELECTRA’s objective function~\cite{clark2020electra} for TABBIE.
These models are finetuned and evaluated for different downstream tasks such as table metadata prediction, table content population, fact-checking, question answering, semantic parsing, and so on. 
For example, TABERT is evaluated on neural semantic parsing; TABBIE for column, row, and column type population tasks; and TUTA is finetuned and evaluated for cell and table type classification tasks. These downstream tasks take either a table or a table with a question as input; they do not consider the inter-table tasks such as dataset discovery, which we consider in this paper. Furthermore, they adopt %Large language models' 
serialization techniques 
from LMs
and represent table values as sentences, whereas we represent 
%the tables efficiently by 
using sketches.

\introparagraph{Dataset Discovery Methods}
Different methods such as keyword search~\cite{2008_cafarella_webtables, 2019_Brickley_google_dataset_search} and table search~\cite{2012_das_finding_related_tables, 2017_lehmberg_stitching_web, 2018_fernandez_aurum} 
have been used for the dataset discovery over data lakes.
Table Union Search (TUS)~\citet{Nargesian2018_Table_union}  presented finding top-k data lake tables that are unionable with a query table. They determine the column unionability using three statistical tests based on value overlap, semantic overlap, and word embedding similarity, and aggregate them to infer table unionability. 
$D^3L$~\citet{2020_bogatu_d3l},  
adopted value overlap and word embedding measures from TUS and added three additional measures to measure column unionability: numerical column distributions, column header similarity, and regular expression matching.
SANTOS\citet{Khatiwada2022_SANTOS} considers the relationships between the column pairs in both query and data lake tables. Fan et el.~\cite{2023_fan_starmie} developed a contrastive-learning-based table union search technique called Starmie that captures the context of the whole table in the form of column embeddings. AutoTUS~\citet{2023_hu_autotus} searches for unionable tables by contextualizing entire tables with embeddings of the relationship between the column pairs. 
Furthermore, there are other table search techniques that focus on finding joinable tables, i.e., given a query table marked with a query column, find the data lake tables that contain a column with overlapping values with the query column. For instance, LSH Ensemble~\citet{2016_zhu_lsh_ensemble} defines joinable table search as a problem of finding data lake tables that have a column having high set containment with the query column. 
They propose a novel index that approximates set containment computation. JOSIE~\citet{2019_zhu_josie} searches for top-k joinable tables based on exact set containment. 
\revision{
AutoJoin~\cite{2017_zhu_auto_join}, on the other hand, determines the fuzzy join between rows of a given pair of joinable tables by employing string transformations such as matching n-grams of cell values and concatenating cell values. 
}
PEXESO~\citet{2021_dong_pexeso_table_discovery} efficiently searches for the top-k semantically joinable tables. Along with a join on exact values, they also consider fuzzy join, i.e., join on abbreviations and synonyms. DeepJoin~\citet{2023_dong_deepjoin} is a deep-learning-based technique that also searches for semantically joinable tables.
WarpGate~\citet{2023_cong_warpgate} is also a system for searching for top-k semantically joinable tables by embedding each column to vector space using pre-trained embeddings. They index column embeddings using Locally Sensitive Hashing~\cite{1999_gionis_similarity_search} for efficient search.
Recently, Chorus~\cite{Kayali_chorus} even demonstrated the ability of Large Language Models to identify joinable columns as a task of completing `pd.merge` command when two tables are presented as pandas dataframe. 
%
%In line with these data discovery methods, 
Our work addresses the task of discovering tables from the data lakes that can be unionable, joinable, and/or subsets of each other. 
Notably, we specialize data discovery neural models based on a single pretrained model.  
% Notably, we propose a single pretrained neural model that can be specialized for any of the above tasks.
The advantage of our approach over 
%Unlike 
above mentioned task specific approaches
%~\cite{2023_fan_starmie, 2023_dong_deepjoin} %our approach 
%such as Starmie, PEXESO, and DeepJoin 
is that, as in natural language processing, much smaller datasets are sufficient
to finetune the model for new data discovery tasks; by 
leveraging the existing tabular knowledge in the pretrained model.

\introparagraph{Benchmarks}
Several tabular benchmarks exist in the literature
~\cite{10.1162/tacl_a_00544, 2023_hulsebos_gittables, 2024_fox_verllm}.
Koutras et al.~\citet{2021_koutras_valentine} created benchmarks for schema matching~\cite{2001_rahm_schema_matching_survey} using open data tables and evaluated the existing matching techniques over them. Mudgal et al~\citet{2018_mudgal_deepmatcher} open-sourced 13 datasets such as the DBLP-Google Scholar author dataset and Walmart-Amazon product datasets for entity matching task~\cite{2016_konda_entity_matching}. 
The WDC Web Table corpus~\cite{2016_lehmberg_wdc_corpus}, containing around 233 million web tables extracted from different web pages,
has been widely used for diverse tasks such as Question Answering~\cite{Herzig2020_TAPAS, 2022_liu_tapex}, Semantic Parsing~\cite{2021_yu_grappa, 2022_liu_tapex, Yin2020_TABERT}, Table Retrieval~\cite{2021_wang_gtr}, Table Metadata Prediction~\cite{Wang2021_TUTA, 2020_deng_turl, 2022_suhara_duduo} and more~\cite{2020_zhang_web_table_survey}. Efthymiou et al.~\citet{2022_efthymiou_semtab22} used knowledge graphs to create tabular benchmarks for column type prediction and column-to-column binary relationship prediction tasks. The VizNet Benchmark~\cite{2019_hu_viznet} has been used to evaluate column type prediction tasks~\cite{2020_zhang_sato, 2019_hulsebos_sherlock}. GitTables~\cite{2023_hulsebos_gittables} is a large table corpus in the literature. However, they contain web tables that are small entity tables with fewer rows and columns than enterprise tables and do not represent enterprise data lake scenarios~\cite{2019_zhu_josie}. 
But the most closely related benchmark to our work is the LakeBench collection of datasets~\cite{lakebench} for identifying relation between pair of tables (Sec.~\ref{finetuning}) Hence, we use that benchmark for finetuning our model.
Further, there are publicly available table search benchmarks that are also relevant to our work which we use for illustrating the use of \ourmethod{} for data discovery. TUS~\cite{Nargesian2018_Table_union} and SANTOS~\cite{Khatiwada2022_SANTOS} are table union search benchmarks which we include in our experiments. At the time of writing this paper, there were no search benchmarks for table join search and table subset search, so we generated \emph{Wiki join search} and \emph{Eurostat subset search} benchmarks (Sec.~\ref{exp:search}). However, recently two new benchmarks have appeared in this space. One is manually annotated table join and union search benchmark~\cite{vldblakebench}  (also called LakeBench\footnote{Not to confuse with Srinivas et al.~\cite{lakebench} LakeBench used for finetuning.}) and another is a semantic table search~\cite{Leventidis_semantic_search} benchmark that used wikipedia categories and navigational links to establish semantic join and union relation. 
%These search benchmarks are, however, not designed for training neural models. 
%To create such benchmarks, we adapt the existing public datasets for pairwise classification and regression tasks.
% We use open and enterprise datasets to create our benchmarks 
%of enterprise data lakes 
% where the tables have 10s of columns and 1000s of rows and their cell values are often numbers or cryptic codes.

%% file: sections/tabsketchfm2.tex
\section{\textsc{\ourmethod{}}}

\label{section:tabsketchfm}

\begin{figure*}
    \centering
    \includegraphics[width=0.8\textwidth]{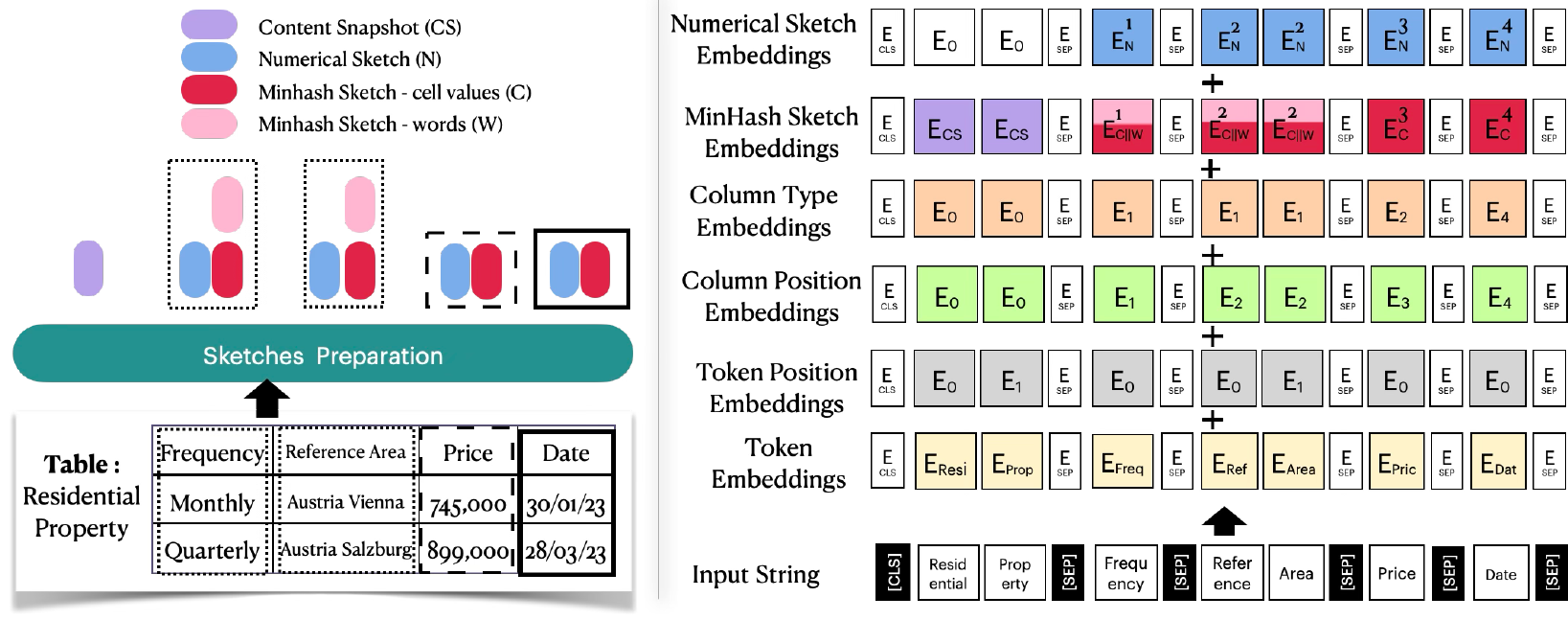}
    \caption{TabSketchFM Table Representation. The left panel depicts the preparation of various sketches for each column of the input table, "Residential Properties". The right panel illustrates the constituent components of the input embedding, which are subsequently summed to form the input embedding for the 12-layered BERT encoder with self-attention mechanism.}
    \label{fig:tabsketchfm}
\end{figure*}

In this section, we detail the architecture of~\ourmethod{}. Like existing tabular models~\cite{Wang2021_TUTA, Herzig2020_TAPAS}, we pretrain the BERT model~\cite{2019_devlin_bert}.
But, unlike them, we adapt the BERT input embeddings to accommodate the numerical inputs.

% \introparagraph{Solution Overview}
% To use TabSketchFM on dataset discovery, we first pre-train a transformer-based model using table sketches as input. After pretraining, we fine-tune the model for downstream tasks like union, join, and subset. The fine-tuned model generates table embeddings from the input sketches, which are then used to create nearest neighbor indexes for search tasks. Next, we explain each step involved in building TabSketchFM.

% Specifically, we prepare numerical sketches (sec.~\ref{sec:sketch_prep}) and construct input embeddings (sec.~\ref{sec:input_repr}).

% The most significant departure from other tabular representation models is in how we represent the table as input in terms of sketches.

\subsection{Sketches Preparation}\label{sec:sketch_prep}

While other models linearize table cells as text, we prepare \textit{sketches}: numerical vectors that capture different features of the table.
We use MinHash signature to convert sets of strings to numerical vectors. MinHash is a technique that uses hash functions to map similar sets to similar hash values, allowing for fast estimation of Jaccard similarity between two sets~\cite{LeskovecRU14_MMD}.\footnote{We use datasketch library~\cite{eric_zhu_2017_290602} to compute MinHash.}
Specifically, we create one table-level sketch called a content snapshot and two types of column-level sketches: numerical sketch and minhash sketches.

\introparagraph{Content Snapshot}  Recognizing that row information could be crucial in detecting 
% the table semantics for data discovery~\cite{Khatiwada2022_SANTOS}, 
similarity of tables, 
we create a sketch from the first 10000 rows.  
We convert each row into a string and generate a MinHash signature from the set of rows. This is the only table-level sketch that we create. 
% \textcolor{orange}{TP: Don't use use a subset of rows? }
%  10000 -- Harsha

    \introparagraph{Numerical Sketches} 
    For each column, we extract a set of statistical features that provide insight into its characteristics. These features include the number of unique values and of nan values, each normalized by the total number of rows. For string columns, we record the average cell width in bytes, as this can impact the likelihood of the column being used as a join key. Notably, very long strings are unlikely to be suitable for joining.  For numeric columns, we calculate additional statistics: percentile values, mean, standard deviation, min and max values. We combine these features into a single vector, which we refer to as a "numerical sketch". To summarize, a numerical sketch vector consists of the following elements: [unique count, NaN count, cell width, 10th percentile, 20th percentile, ..., 90th percentile, mean, standard deviation, min value, max value]. When possible, we convert date columns to timestamps and treat them as numeric columns.

    \introparagraph{MinHash Sketches}  For each column, we treat cell-value as a string and compute a MinHash signature from the set of cell-values.\footnote{
    While it may not make sense to convert float or integer columns into MinHash signatures, we do so because it is often difficult to tell if a column is truly a float or an integer or is really a categorical value.} For string columns, we also compute a MinHash signature for set of words within the column. The rationale for this second MinHash is that the words 
    can sometimes capture its semantics (e.g., if \texttt{street} appears in two different columns, they maybe both about address, which is useful if one 
    considers a union of two tables that do not necessarily share cell values). Consider the Reference Area column in the table shown in~\cref{fig:tabsketchfm}. The cell-values MinHash Sketch treats "Austria Vienna" as an element in the set of strings, whereas "Austria" and "Vienna" are treated as separate strings in words-based MinHash signature.
    Both MinHashes are concatenated
    into a single input vector for string columns.  For numerical and date columns, only the MinHash for the cell values is included in the input vector.

The left side of Figure ~\ref{fig:tabsketchfm} illustrates the process of preparing the content snapshot for the table, as well as generating numerical and MinHash sketches for each individual column. 

\subsection{Input Representation}
\label{sec:input_repr}
Now that we have these sketches, which are arrays of numbers (vectors),
a key question is how to adapt BERT's architecture to consume these numeric values. Transformer models construct input embeddings from any given sentence by summing three different embeddings: 1. token embeddings---embedding of the input sentence broken into tokens, 2. token type embedding---that indicates if the token belongs to the first sentence or second, and 3. position embeddings---that indicates the position of the token in the input sentence. We adapt this to table-specific inputs in the following manner. We first generate an ``input string'' that is concatenation of all the table metadata and column names delimited by the separator token. The input string for table in Fig~\ref{fig:tabsketchfm} is shown at the bottom right. The input embedding is constructed from this input string by summing the following different embeddings:
    
    \introparagraph{1. Token Embeddings} Similar to the BERT, the first embedding used for the input string is the token embeddings. We initialize the embedding weights from the pretrained BERT uncased model to leverage the extensive pretraining of BERT for its vocabulary. This is represented in Fig~\ref{fig:tabsketchfm} with yellow boxes.
    
    \introparagraph{2. Token Position Embeddings} 
    Considering each column analogous to a sentence in a paragraph, we re-purpose token positional embeddings to reflect a token's position within a column name. For instance, as shown in~\cref{fig:tabsketchfm} with light gray boxes, the token position of the token \texttt{Area} is 1 because it is the second token in that column (after \texttt{Reference}).
    %\item 
    
    \introparagraph{3. Column Position Embeddings}
    Columns themselves have positions and are encoded through a new embedding layer which can range from 1 to the total number of columns,\footnote{Position 0 is reserved for any tokens in the table description.} as shown in green boxes in~\cref{fig:tabsketchfm}. The rationale for including column positions is that they sometimes do have semantics; e.g., a street address next to the city, followed by a zip code.  Of course, table semantics do not change as a function of column order; nevertheless, we included a position in case it helps the model understand column semantics, with the assumption that the attentional mechanism would disregard order if necessary.
    %\item 
    
    \introparagraph{4. Column Type Embeddings} The data types of columns: string, date, integer, or float are encoded through additional embedding. In order to determine column type, we made a best-case effort to parse the first 10 values of each column as dates, integers, or floats and defaulted to string if we could not convert them.  In mixed-type columns, this can yield poor results, but at the very worst, at least one of the types was assigned to these columns.  This is shown in~\cref{fig:tabsketchfm} in orange, where string, integer, float and date types are represented with integers 1, 2, 3, and 4, resp.
    %\item 
    
    \introparagraph{5. MinHash Sketch Embeddings} This embedding encapsulates all the MinHash sketches, including those derived from cell values, words, and content snapshots. The embedding is obtained by passing the MinHash vectors through a linear transformation layer. Notably, the MinHash embedding for table meta-data tokens corresponds to the content snapshot embedding, denoted as $E_{CS}$, which is visually represented by purple boxes. For numerical columns, the MinHash embedding is the embedding of the MinHash signature derived from the set of corresponding cell values, denoted as $E_C$, and illustrated in red. In contrast, the MinHash embedding for string columns is constructed by concatenating the MinHash signatures from both the corresponding set of  cell values and the set of words, resulting in the embedding $E_{C||W}$, shown with red and pink.

    \introparagraph{6. Numerical Sketch Embedding} Lastly, each column's numerical sketch vector is fed into a linear transformation layer, and the output is utilized as the numerical sketch embedding.
%\end{itemize} 

The sum of these embeddings is used as the input embedding for the 12-layered BERT encoder model with self-attention from HuggingFace\footnote{\url{https://huggingface.co/docs/transformers/model_doc/bert}}. The self-attention in BERT is bi-directional: each token can attend to the tokens on both its left and the right side. This bi-directional attention helps the TabSketchFM model to disambiguate column names like \texttt{Age} which can mean mean person's age in an \texttt{Employee} table vs age of a building in a \texttt{Residential Properties} table.

%% file: sections/pretraining.tex
\begin{figure}[!ht]
   \begin{subfigure}[b]{0.45\columnwidth}
       \centering
       \includegraphics[width=0.7\textwidth]{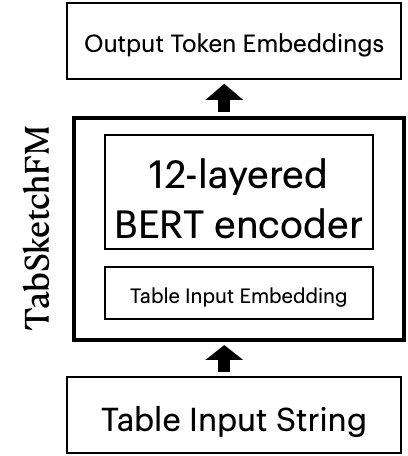}
       \caption{TabSketchFM Pretraining.}
        \label{fig:pretraining_arch}
   \end{subfigure}
   \hfill
   \begin{subfigure}[b]{0.45\columnwidth}
       \centering
       \includegraphics[width=0.7\textwidth]{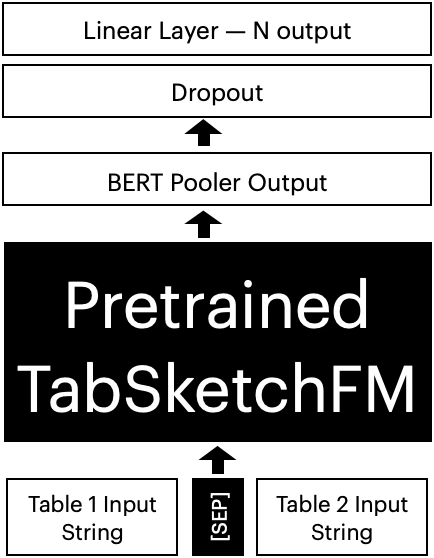}
       \caption{TabSketchFM Finetuning.}
       \label{fig:finetuning_arch}
   \end{subfigure}
   \caption{\revision{(a) For pretraining, the table input embedding (see~\cref{sec:input_repr}) is passed through 12-layered best encoder to generate output token embeddings. The output token embeddings are used for MLM objective with cross entropy loss(~\cref{sec:pretraining}). We initialize the weight of the BERT encoder as well as the token embeddings from the pretrained base uncased BERT model. (b)  Two input tables are concatenated and passed through the  pretrained TabSketchFM. The BERT pooler output (embedding of the first token) is taken and passed through a dropout and a linear layer to generate output of size N.}
}
   \label{fig:tabsketchfm_pre_fine}
\end{figure}

\subsection{Pretraining}
\label{sec:pretraining}
%Now we explain how we pretrain \ourmethod{} using the architecture described in ~\cref{section:tabsketchfm}. 

\introparagraph{Dataset}
Existing works generally use non-public data~\cite{10.1162/tacl_a_00544} or web tables~\cite{2016_lehmberg_wdc_corpus} to pretrain the models.
Unlike enterprise data, web tables often have few rows and columns to ease human consumption~\cite{2008_cafarella_webtables} and they focus on entities popular on the web (e.g., countries). Their performance may not generalize to enterprise data, where the tables have a large number of rows, the entities are often domain-specific, contain cryptic code words, and have a lot of numerical information.
Therefore, we created a de-duplicated pretraining dataset of $197,254$ enterprise-like tables (CSVs) from CKAN and Socrata that are legally cleared to have the 
open 
%correct
%ak: correct license sounds vague to me
licenses.  
%We used this data instead of web tables \cite{2016_lehmberg_wdc_corpus} to build on tables that are more enterprise like. 
% ~\Cref{tab:pretraining_cardinality} shows the average numbers of columns and rows in the tables which are in the order of tens and thousands respectively.  
%
%as shown in ~\cref{tab:pretraining_cardinality},
This pretrain dataset contains $2234.5$ rows and $35.8$ columns in each table on average. About $66\%$ of columns were non-string, resembling an enterprise datalake.

\iffalse 
\begin{table}
\small
%\fontsize{8}{10}\selectfont
\setlength{\tabcolsep}{3pt}
\centering
\caption{Cardinality and data type distribution of the Opendata pretraining dataset.}
  \label{tab:pretraining_cardinality}
  \begin{tabular}{lrrrrrr}
    \toprule
    \# Tables&Avg. Rows&Avg. Cols &String&Float&Integer&Date\\
    \midrule
 197,254 & 2234.5 & 35.8  & 2,430,684 & 3,768,508 & 630,976 & 231,664\\
  \bottomrule
\end{tabular}
\end{table}

\fi

% %\begin{wraptable}{r}{18em}
% \begin{table}
% \small
% %\fontsize{8}{10}\selectfont
% \setlength{\tabcolsep}{3pt}
% \centering
% \caption{Pretraining data type distribution for columns.}
%   \label{tab:pretraining_data_types}
%   \begin{tabular}{lrrrr}
%     \toprule
%     Benchmark&String&Float&Integer&Date\\
%     \midrule
%     Opendata    & 2,430,684 & 3,768,508 & 630,976 & 231,664 \\
%   \bottomrule
% \end{tabular}
% \end{table}

\introparagraph{Data Augmentation}
In the image processing literature~\cite{2006_acharya_image_processing_book}, different transformations are applied over an original image to generate new image samples for training. This makes the trained model robust on different variants of the same image.
We want the model to be robust to the order of columns such that shuffling columns in a table does not impact its semantics. So, we created three different versions of the table, by changing the column order, which in turn, changed the table's content snapshot~(see \cref{section:tabsketchfm}). 
%The idea here is inspired by the augmentations performed on images in neural networks literature, where a given input image is cropped, rotated etc. to make the neural network more robust to different variants of the same image.
This increases the total number of pretraining tables to 290,948 and for each of them, we create signatures as explained in~\cref{section:tabsketchfm}.

%For each table in Figure \ref{tab:pretraining_cardinality}, we created 3 different versions of the table, by changing column order, which also changed the content snapshot of the table.  The idea here is similar to the sorts of augmentations performed in the image neural networks literature, where a given input image is cropped, rotated or more generally transformed in some way to maximize the use of data.  This resulted in 290,948 tables for which we created signatures.

\introparagraph{Method} We employ the standard Mask Language Modeling (MLM) as the pretraining objective for our model, as in most text-based language models. %MLM is famous for Text-based LLMs that are trained to understand the context in a given input sequence. 
First, a few tokens are randomly sampled according to the masked language probability. Then they are substituted with a [MASK] token to create an input sequence. The input sequence is fed to the model, and the model is asked to predict the substituted words. This can be viewed as a classification task where the original word is the label and the model's entire vocabulary is the set of possible labels. Cross-entropy loss is computed for each predicted word in the input text that was masked. 
%The neighboring columns' contexts could be helpful in understanding the context of a column in a table. 
%It is important to understand the neighboring columns in a table in order to understand its semantics. 
%For good performance, a tabular model can exploit the context of the entire table such as the neighboring columns of a given column. 
%\aamod{The sentence above is not connected anywhere in the text.}
%
%
%
\begin{equation}
   \label{eq:cross_entropyloss}
    \frac{-1}{N} (\sum_{n=1}^{N} y_{i} * log(\hat{y_{i}}))
\end{equation}
\Cref{eq:cross_entropyloss} defines cross-entropy loss, where $y_{i}$ is the class labels in the ground truth, $\hat{y_{i}}$ is the predicted class label and N is the number of training examples in the train set.
%which corresponds to the vocabulary size.

\begin{figure}
    \centering
    \includegraphics[width=1\columnwidth]{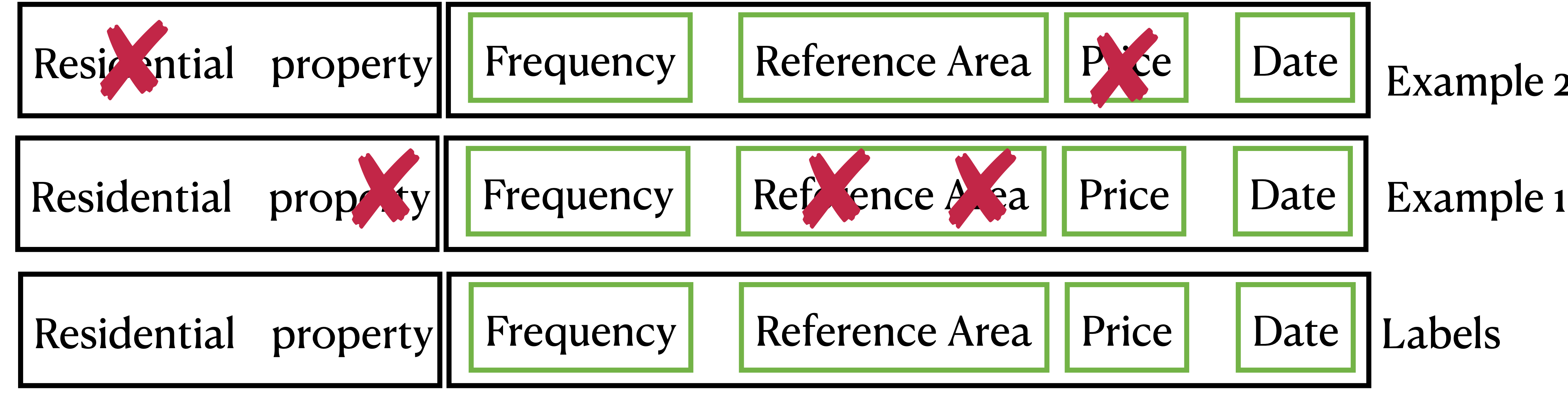}
    \caption{A combination of whole column masking, and MLM probability masking to generate up to 5 examples. The box on the left hand side is the table description and the one on the right hand side is the column names}
    \vspace{-2em}
    \label{fig:masking}
\end{figure}

In our scenario, for each table, we mask a single column name such that all tokens corresponding to that column are masked. This is analogous to the natural language literature's whole word masking where all tokens corresponding to a word are masked. 
In addition, we used MLM probability to mask the tokens in the table description as well.  
As shown in~\cref{fig:masking}, we generate several samples from a single table. Note that each masking gives us an example. 
For small tables with less than $5$ columns, we mask each of its columns one after another.
%was masked across the examples.
%To avoid over-representation of a table, 
%For tables with more than 5 columns, 
However, if a table has a large number of columns, this produces a lot of examples for the same table, over-representing it.
So, for large tables with more than $5$ columns, we randomly select five columns and mask them.
%5 different training examples by choosing the column to mask at random.
%for the each example derived from the same table.
%This ensures that there was not an over-representation of tables with a large number of columns in the data. 
Following this strategy over the augmented pretraining dataset, we get $730,553$ examples in training, $54,430$ examples in validation, and $58,141$ examples in test sets. We provide the experimental setup for pretraining in~\cref{sec:exp}. \revision{Figure~\ref{fig:pretraining_arch} shows the TabSketchFM pretraining architecture.}

%\aamod{I think we should provide details like number of iteration, parameters, training time, etc.}

%% file: sections/finetuning.tex
\subsection{Fine-tuning}\label{finetuning}

To use our neural tabular model for identifying relevant tables in the data lakes, we finetune \ourmethod{} on the \emph{LakeBench} collection~\cite{lakebench}. LakeBench is a collection of 8 benchmark that includes data from various sources such as open government data from CKAN and Socrata, economic data from the European Central Bank~\cite{eurosystem}, Spider~\cite{yu-etal-2018-spider}, and synthesized data from large knowledge graphs such as Wikidata~\cite{10.1145/2629489}.  The benchmarks cover binary classification, regression, and multi-label classification for the tasks of identifying table unionability, table joinability, and table subset. It contains 3 datasets for identifying unionable tables---TUS-SANTOS, Wiki Union, and ECB Union; 4 datasets for identifying joinable tables---Wiki Jaccard, Wiki Containment, Spider-OpenData, and ECB Join; and one dataset for identifying table subsets---CKAN Subset. \revision{Of these datasets, Wiki Jaccard, Wiki Containment and ECB Union are formulated as regression task {to estimate jaccard similarity value, containment value, and the number of unionable columns, resp. (cf ~\cite{lakebench})}; ECB Join is  multi-class classification {to predicts the joinable columns in the table}; and remaining are classification tasks.}  Table~\ref{tab:benchmark_cardinality} summarizes the main characteristics of the datasets to show the type of tasks we used for fine tuning. 

\begin{table*} 
\small
%\fontsize{8}{10}\selectfont
\setlength{\tabcolsep}{5pt}
\centering
\caption{Cardinality of all the datasets in LakeBench, as well as search benchmarks in this paper.}
  \label{tab:benchmark_cardinality}
  \resizebox{0.8\textwidth}{!}{
  \begin{tabular}{llrrr|rrr|rrrr}
    \toprule
    Benchmark&Task&\# Tables&Avg. Rows&Avg. Cols& \multicolumn{3}{ c }{\# Table Pairs}& \multicolumn{4}{ |c }{Data type distribution (\%)}\\
    &&&&&Train&Test&Valid & String&Int.&Float&Date\\
    \midrule
    TUS-SANTOS & Binary Classification & 1127 & 4285.17 & 13.04 & 16592 & 3566 & 3552  & 77.94 & 8.62 & 7.51 & 5.93 \\ 
    % 1078 & 4280.46 & 13.02 & 16592 &  3566 &  3552 \\
    Wiki Union & Binary Classification & 40752 & 51.05 & 2.66 & 301108 & 37638 & 37638  & 57.97 & 14.38 & 24.25 & 3.4 \\
    % 36034 & 53.46 & 2.61 & 301108 &  37638 & 37638 \\ 
    ECB Union & Regression & 4226 & 292.47 & 36.3 & 15344 & 1910 & 1906  & 47.72 & 14.05 & 36.31 & 1.92 \\
    % 3685 & 285.49 & 36.26 & 15344 &  1910 &  1906 \\
    \midrule
    Wiki Jaccard & Regression & 8489 & 47.3 & 2.8 & 12703 & 1412 & 1572 & 57.5 & 15.66 & 19.76 & 7.07 \\
    % 5532 & 46.92 & 2.81 & 12703 &  1412 & 1572\\
    Wiki Containment & Regression & 10318 & 47.15 & 2.79 & 21007 & 2343 & 2593 & 57.26 & 15.26 & 20.58 & 6.9 \\
    % 6976 & 46.75 & 2.8 & 21007 &  2343 & 2593\\
    Spider-OpenData & Binary Classification & 10730 & 1208.87 & 9.09 & 5146 & 742 & 1474  & 42.22 & 18.51 & 32.62 & 6.66\\
    % 5365 & 1208.9 & 8.93 & 5146 &  742 & 1474 \\
    ECB Join&Multi-label Clasification & 74 & 8772.24 & 34.47 & 1780 & 222 & 223 & 52.14 & 7.8 & 37.79 & 2.27\\
    % 72 & 8738.14 & 34.39 & 1780 &  222 &  223) \\
    %Subset Fred&Binary Classification & 32,575 & 165.94 & 2.00 & 0.80 & 0.10 & 0.10\\
    \midrule
    CKAN Subset & Binary Classification & 36545 & 1832.58 & 25.37 & 24562 & 2993 & 3010 & 31.75 & 17.53 & 46.14 & 4.58\\
    \midrule
    Eurostat Subset & Search & 38904 & 2157 & 10.46 & & & & 64.63 & 9.03 & 7.83 & 18.50 \\
    Wikijoin & Search & 46521 & 49.64 & 2.68 & & & & 58.13 & 13.35 & 25.0 & 3.50 \\
    % 24086 & 1813.54 & 26.0 & 24562 &  2993 &  3010 \\
    
  \bottomrule
\end{tabular}
}
\end{table*}

For each finetuning task, we create 
a cross-encoder~\cite{2022_du_cross_encoder}, 
which evaluates if a pair of tables can be joined, unioned or are subsets of each other. \revision{Figure~\ref{fig:finetuning_arch} shows the architecture of the TabSketchFM cross-encoder used for finetuning.  For the classification and regression tasks, we use the BERT pooler output, that is the embedding of the first [CLS] token and pass it through a dropout and a linear layer to generate output of size N. For binary classification tasks, the output layer has N=2, and we use cross-entropy loss. For regression tasks, the output layer has N=1 and we use mean-squared error loss. For multi-class classification tasks, the output layer has N = number of classes and we use binary cross-entropy with logit loss for finetuning.}
% We use cross-entropy loss for classification tasks, mean squared error for regression tasks, and binary cross-entropy with logit loss for multi-class classification tasks.  
The goal of building multiple cross encoders for the same task across different datasets was to examine whether the cross encoders would generalize across datasets; allowing re-use to new data with no additional training.

\revision{
\subsection{TabSketchFM Overview}\label{sec:overview}
We propose TabSketchFM a novel sketch-based tabular representation learning architecture for dataset discovery tasks. Section~\ref{sec:input_repr} explains how we generate the sketches, and Section~\ref{sec:input_repr} describes how we adapt BERT embedding module to incorporate the numerical sketches. The model is pretrained on opendata using MLM objective to predict the columns with cross-entropy loss, described in Section~\ref{sec:pretraining}. After pretraining, we fine-tune the model for downstream tasks like union, join, and subset, as elaborated in Sec.~\ref{finetuning}. Finally, we use the finetuned TabSketchFM for data discovery tasks, in the experiments. To do so, we extract the table embeddings from the finetuned TabSketchFM, and use that to create nearest neighbor indexes for search tasks. We evaluate all these aspects in the next section. 
}

%% file: sections/experiments.tex
\section{Experiments}\label{sec:exp}

Our empirical work is organized around the following research questions:
%This section is organized around three research questions 
\begin{enumerate}[start=1,label={ \bfseries Q\arabic*.},ref={Q\arabic*}]
    \item How does sketch-based \ourmethod{} compare to existing pretrained tabular models %for building fine tuned models for data discovery tasks?
    when finetuned for relevant table identification tasks in LakeBench?\label{Q1}
    \item How do different sketches of \ourmethod{} impact different tasks?\label{Q2}
    \item How well do these fine tuned \ourmethod{} models perform on data discovery tasks?\label{Q3}
    \item Can \ourmethod{} fine tuned on a single task and domain, adapt and generalize to a new domain?\label{Q4}
\end{enumerate}

% \hk{Main section doesnt use datasketch term. Only Sketch}

\begin{table*}[t]
    \caption{Performance (avg $\pm$ stdev) of TabSketchFM vs. other baseline models on LakeBench averaged across five random seeds. \textbf{Best} performance is highlighted in bold, and the second best is \underline{underlined}.}
    \label{lakebench_results}
    \begin{center}
\resizebox{0.8\textwidth}{!}{
    \begin{tabular}{lc|cc|ccc}
        \toprule
        Tasks                 & Vanilla BERT     & TAPAS          & TABBIE      & TUTA                              & TaBERT      & \ourmethod{} (ours) \\
        \midrule
        TUS-SANTOS (F1)           & $\boldsymbol{0.99 \pm 0.00}$ & $0.34 \pm 0.00$    & $\underline{0.75 \pm 0.22}$  & $\boldsymbol{0.99 \pm 0.00}$  & $\boldsymbol{0.99 \pm 0.00}$       & $\boldsymbol{0.99  \pm  0.00}$       \\
        Wiki Union (F1)       & $0.33 \pm 0.00$ & $0.41 \pm 0.00$  & $0.64 \pm 0.12$  & $0.33 \pm 0.00$    & $ \boldsymbol{0.97 \pm 0.00}$ & $\underline{0.94  \pm  0.00}$      \\
        ECB Union (R2)        & $0.03 \pm 0.03 $ & $-0.01 \pm 0.01$ & $0.02 \pm 0.01$ & $\underline{0.87 \pm 0.01}$  & $0.35 \pm 0.06$ & $\boldsymbol{0.90 \pm  0.03}$     \\
        \midrule
        Wiki Jaccard (R2)     & $0.00 \pm 0.00$ & $-0.03 \pm 0.03$ &  $0.25 \pm 0.02$ & $\underline{0.43 \pm 0.02}$   & $0.33 \pm 0.03$ & $ \boldsymbol{0.58  \pm  0.03}$      \\
        Wiki Containment (R2) & $0.00 \pm 0.00$ & $0.00 \pm 0.00$ & $0.21 \pm 0.01$ & $\underline{0.35 \pm 0.03}$ &  $0.30 \pm 0.03$ & $ \boldsymbol{0.58  \pm  0.02}$       \\
        \spiderlb{} (F1)           & $0.71 \pm 0.06$ & $0.65 \pm 0.00$ & $0.57 \pm 0.00$ & $0.76 \pm 0.12$ &  $ \boldsymbol{0.87 \pm 0.02}$ & $\underline{0.83  \pm  0.01}$      \\
        ECB Join (F1)         & $0.63 \pm 0.04$ & $0.40 \pm 0.00$ & $0.42 \pm 0.03$ & $\underline{0.81 \pm 0.01}$   &  $0.79 \pm 0.03$ & $ \boldsymbol{0.86  \pm  0.01}$       \\
        \midrule
        CKAN Subset (F1)      & $0.43 \pm 0.00$ & $0.43 \pm 0.00$ & $0.43 \pm 0.00$    & $0.43 \pm 0.00$ & $\underline{0.43 \pm 0.00}$   & $ \boldsymbol{0.98  \pm  0.00}$  \\
        \bottomrule
    \end{tabular}
            }
    \end{center}
\end{table*}

% \subsection{Experimental Setup}
% \hk{I want our subsection numbers to match the question numbers}

% We run all our experiments using Python $3.9$.
% \hk{This can go in appendix or code readme.}

We pretrain \ourmethod{} using $4$ A100 $40$GB GPUs for $2$ days, until the model converged.\footnote{The source code is available on \url{https://github.com/ibm/tabsketchfm}} Our pretrained model contains $118M$ parameters, similar to other row-based models in our experiments. We use patience of $5$, i.e., we consider the model as having converged if the validation loss does not decrease for more than $5$ epochs. As studied in the literature~\cite{2020_zhou_patience} this not only reduces training time but also helps us avoid overfitting.
For finetuning the cross encoders, for all tasks, we use one A100 $40$GB GPU. Most finetuning tasks converged within $6$ hours using the same 5 epoch patience as pretraining except for the Wiki Union benchmark which took 24 hours. Our code is accessible at \url{https://github.com/ibm/tabsketchfm}.
\subsection{Fine-tuning models on \lakebench{} Tasks}
\label{exp:finetuning}
We evaluate our finetuned cross-encoders and the existing value-based tabular foundational models for the LakeBench unionability, joinability, and subset tasks.

\subsubsection{Baselines}
We compare \ourmethod{} against \emph{four} publicly available tabular foundational models: TUTA, TaBERT, TABBIE, and TAPAS.  
We adapted these models for Lakebench data discovery tasks by building a dual encoder architecture. Each encoder represents the pretrained model with shared parameters; and the same model is used to encode each element of a table pair.  The embeddings from the last layer of the encoders were concatenated and passed through a two-layered MLP (multi-layered perceptron).  We used this architecture as opposed to the cross-encoder architecture we used for \ourmethod{} because in general it is unclear how to feed two different tables with different row sizes into a cross-encoder.  It gets even more complicated when one considers models such as TUTA that model hierarchical information in the table as a tree.  While we were able to finetune the TUTA and TaBERT code in this dual encoder architecture, the code of TAPAS and TABBIE were less amenable to finetuning. Hence, for TAPAS and TABBIE, we froze their pretrained models while finetuning, but allowed the two layers above the model to learn the weights.  In this case, the pretrained model can be seen as producing an embedding that is then fed into two layered network that tunes for a specific task. We will make the code for all the baselines available.
% We provide the architecture details and hidden layer sizes in the appendix. 
Below, we provide details on 
%quick overview of 
%ak: better to say we are comprehensive
model-specific implementation issues.
 
\begin{itemize}
    \item \textbf{TaBERT} provides two types of embeddings for a table: context embeddings corresponding to the tokens in the context, and column embeddings corresponding to each column in the table~\cite{Yin2020_TABERT}. For each table, we compute both embeddings for top $10,000$ rows. In \citet{Yin2020_TABERT}, a content snapshot approach is described which creates synthetic rows based on the n-gram matches of the input question with the cell values. We could not experiment with this feature, since 1. this technique is not part of the authors' provided source code\footnote{\url{https://github.com/facebookresearch/TaBERT}} and 2. LakeBench tasks do not contain a question. So, we compute these embeddings for each table, mean-pooled over the context tokens and tabular columns respectively. The final embedding for the table pair is computed as the concatenation of context and column embeddings obtained for each table. 
    \item \textbf{TUTA} creates a tree-based attention matrix over all the tokens in the table, which has significant memory requirements. To overcome this, we obtain the top $256$ rows and columns of the table, and the first $256$ tokens of the table sequence. 
   \item \textbf{TAPAS} requires a natural language query and a table as input~\cite{Herzig2020_TAPAS}. For our dataset discovery tasks, we sent an empty string as a natural language query and use $512$ serialized row based version of the table as the embedding. 
    \item \textbf{TABBIE} provides row and column embeddings of the table~\cite{Iida2021_TABBIE}. For our work, we obtained the row embeddings for each row in the table. Following the original work, we use the first 30 rows and 20 columns of the table. These row embeddings are combined using the mean operation and the resulting vector is frozen as table embedding.
\end{itemize}

We also added a naive baseline, \textbf{Vanilla BERT}, to examine to what extent 
%data discovery tasks on 
LakeBench tasks can be performed based on column headers alone; this provides a measure of the difficulty of a given task.  For \emph{Vanilla BERT}, column headers were provided for the two tables as two sentences, and the BERT model was finetuned on the specific LakeBench task as were other models.  

% In some cases, column names were deliberately masked out to ensure that the neural model could not rely on column names alone.  Nevertheless, other factors such as the differences in number of columns for positive pairs versus negative pairs may also clue systems.  Therefore our naive baseline was a vanilla pretrained model designed to ensure that the test could not be performed trivially. 
%  \hk{Need more description of Vanilla Bert here.}

% if the column is numeric, then we represent it using the 10th, 20th.. 90th quantiles of the distribution. Whereas, string and date columns are encoded using their minhash values and the number of unique values. For all the columns types, we compute the number of NaN values. For the model to understand how the actual values in the table, we sample 10,000 rows and compute its minhash. This content snapshot, is also fed to the model. 

% For each table, the model concatenates all the column representations and creates 3 arrays, (1) numeric values: one for the numeric values which would include the number of nans, number of unique and the quantile sketch of a column if applicable. (2) minhash: this includes the minhash of all the string/date columns and the content snapshot 

\subsubsection{Results}

\Cref{lakebench_results} compares the performance of \ourmethod{} with baseline models on LakeBench. For regression tasks, we report R2 statistics, and for (binary and multiclass) classification tasks, we report a weighted F1 score to handle skew in classes\footnote{Our metrics were implemented from \texttt{scikit-learn}: \url{https://scikit-learn.org/stable/modules/model_evaluation.html}}. The best performance is highlighted in bold and the second best is underlined. All the results presented are aggregated over 5 random seeds. 
% \revision{
We see that on \emph{five of the eight tasks, {\ourmethod{}} outperformed all baselines, performed as well as the best methods on one, and performed second best on the remaining two}.
% }

%We include other statistics, like mean and R values in the supplementary.
%
The binary classification adaptation of the TUS-SANTOS benchmark, was too easy and could be solved on the basis of column headers alone\footnote{We added column headers for this benchmark, unlike in the literature, but removing headers made little difference to the results of tabular models.}, as evident from the performance of the Vanilla BERT model.  Nevertheless, TAPAS and TABBIE perform relatively poorly on it, pointing to the importance of including the Vanilla BERT as a baseline. 
%
%
% row based models. 

\begin{figure*}
\begin{minipage}[b]{\columnwidth}
\captionof{table}{Importance of different sketches on \emph{\ourmethod{}} performance for the different tasks.  Performance of sketches that are equivalent (within 2 points) to overall performance is highlighted in \textcolor{darkgreen}{green}. TUS-SANTOS is not considered because it can be performed based on column headers alone.}
    \label{tab:ablation_results1}
    \resizebox{\columnwidth}{!}{
    \begin{tabular}{l|rrr|r}
    \toprule
    & \multicolumn{3}{|c|}{Using only} & \\
    Tasks  &  {\small MinHash}  & {\small Numerical}	&	{\small Content} & {\small \emph{\ourmethod{}} } \\
      &  {\small sketches}  & {\small sketches}	&	{\small snapshot} & {\small (w/ everything)}
    \\
   
    \midrule
        %{\small TUS-SANTOS (F1)} & 0.989 &	0.990	&	0.994 & 0.993 \\
        {\small Wiki Union (F1)}  & 0.914 & 0.804	& 0.897	&	0.940 \\
        {\small ECB Union (R2)} & 0.829	&	0.498	&	0.752	&	0.897   \\
        \midrule
        {\small Wiki Jacc. (R2)} & 0.537	&	0.318	&	0.314	&	0.577  \\
        {\small Wiki Cont. (R2)} & \textcolor{darkgreen}{0.628}	&	0.252	&	0.301	&	0.587   \\
        {\small Spider-Op. (F1) }&  \textcolor{darkgreen}{0.831}&	0.817	&	0.797	&	0.831  \\
        {\small ECB Join (F1) }&  \textcolor{darkgreen}{0.874} &	0.812		& 0.815 &	0.856    \\
        \midrule
        {\small CKAN Subset (F1)} &  0.431	&	\textcolor{darkgreen}{0.984}	&	0.431	&	0.986 \\
        \bottomrule
    \end{tabular}
    }
\end{minipage}
\hfill
\begin{minipage}[b]{\columnwidth}
\captionof{table}{Removing a sketch to study the impact of it on \emph{\ourmethod{}} performance for the different tasks.  Performance of sketches that declined more than 2 points compared to overall performance is highlighted in {\color{red}red}.}
\label{tab:ablation_results2}
\resizebox{\columnwidth}{!}{
\begin{tabular}{l|rrr|r}
\toprule
& \multicolumn{3}{|c|}{Removing only} & \\
Tasks  &  {\small MinHash}  & {\small Numerical}	&	{\small Content} & {\small \emph{\ourmethod{}} } \\
&  {\small sketches}  & {\small sketches}	&	{\small snapshot} & {\small (w/ everything)}\\
\midrule
%{\small TUS-SANTOS (F1) }& 0.992 &	0.992 &	0.989 & 0.993 \\
{\small Wiki Union (F1) } & 0.933 &		0.927	&	0.931 & 0.940 \\
{\small ECB Union (R2) }& 0.770	&	{\color{red}0.872} &	0.897  & 0.897 \\
\midrule
{\small Wiki Jacc. (R2)} & {\color{red}0.425} &		0.565	&	{\color{red}0.519} & 0.577  \\
{\small Wiki Cont. (R2)} & {\color{red}0.358} &		0.598	&	{\color{red}0.559} & 0.586\\
{\small Spider-Op. (F1)} &  0.814 &	 0.851 &	0.847 & 0.831  \\
{\small ECB Join (F1) }&  {\color{red}0.812}	&	0.855	&	0.846 & 0.855   \\
\midrule
{\small CKAN Subset (F1)} & {\color{red}0.431} &	{\color{red}0.431} & 0.980 & 0.986 \\
\bottomrule
\end{tabular}
}
\end{minipage}
\end{figure*}

On the CKAN Subset benchmark, we find that the performance of most models was comparable to random guessing since the column headers were exactly the same, and most systems (except TaBERT) did not have a view of the entire dataset.  On Wiki Union and on \spiderlb, TaBERT outperformed {\ourmethod{}}.  Even so, {\ourmethod{}} achieved the second best performance on these two tasks. 

We also performed an error analysis on table pairs that {\ourmethod{}} got wrong on Wiki-Union, Spider open data and CKAN.  In all these cases, mean hamming distances of MinHash sketches for positive pairs were equivalent to those from negative pairs; suggesting that the sketches alone did not contain sufficient information to discriminate those examples.  TaBERT's superior performance on the Wiki Union case, combined with the fact that lack of discrimination of MinHashes provides a clue to at least one weakness in the current version of \ourmethod{}, which is that we do not consider any values from the column at all.  As shown in~\Cref{fig:wiki_examples}, two columns may be unionable when their values map to the same domain (e.g., different municipalities in Slovakia), with minimal overlap in actual values.  In such cases, a system such as TaBERT can still determine that the cell values are similar, and belong to the same semantic type, whereas \ourmethod{} cannot.  We show how to incorporate cell values externally in the section on search which addresses this issue. 

\textbf{In summary}, the answer to ~\ref{Q1} is that sketches-based \ourmethod{} is in general superior to other pre-trained models for data discovery in LakeBench. Note that this evaluation does not directly speak to performance on search tasks in data discovery; but it does evaluate the usefulness of sketches for building a good model tuned for data discovery. 
 We turn next to investigating how different sketch types help specific tasks, before returning to the question of how well these fine tuned models perform on search.

\subsection{Effectiveness of sketch types on task types}\label{exp:ablation}

To explore the relevance of the 
%numerical sketches, MinHash 
sketches and content snapshots for the different data discovery tasks, we 
conduct an ablation study where we use only one type of sketch.
%used that sketch alone and zeroed out the values of the other sketches. 
% We then evaluated it on the same \lakebench{} tasks with the random seed 0; 
The results on \lakebench{} tasks with the random seed 0
are shown in~\cref{tab:ablation_results1}. 
In ~\cref{tab:ablation_results2} we present another ablation study where we remove
one sketch and retained the rest. 
Not surprisingly, MinHash sketches are crucial for join tasks, where MinHash alone seems to dominate overall performance of \emph{\ourmethod{}}.  Subset selection clearly relies on the overlap of the data distribution; numerical sketches alone match the performance of \emph{\ourmethod{}}.  Given that almost 69\% of the columns were numerical in the subset benchmark, this outcome might be expected. The effect of content snapshot was less clear. Looking at~\cref{tab:ablation_results2}, it is evident that the removal of content snapshot does in fact contribute somewhat to the Wiki join tasks.  Numerical sketches contribute to ECB union which is perhaps because approximately 50\% of the columns are not strings.  Removal of MinHashes mirrors the result from earlier, which is, join tasks are most affected. Surprisingly, the removal of MinHashes affected the subset task; even though numerical sketches by themselves produced good performance. It is possible that there are complex interactions between the sketches.
%
%but this is something we leave for future work. 
%ak: I think what we said is already enough, we should not leave a lot for future, they will otherwise ask during revision :) 
\textbf{In summary,} we answer~\ref{Q2} from our ablation studies that different sketches play different roles across tasks, and may even interact to determining data discovery performance.

\input{sections/search_experiments_v2}

%% file: sections/search_experiments_v2.tex
\subsection{Application to search}\label{exp:search}

We evaluate the embeddings from our finetuned~\ourmethod{} for three data discovery tasks: join search, union search, and subset search.

\begin{figure*}
\begin{subfigure}{\columnwidth}
    \centering
    \includegraphics[width=0.8\columnwidth]{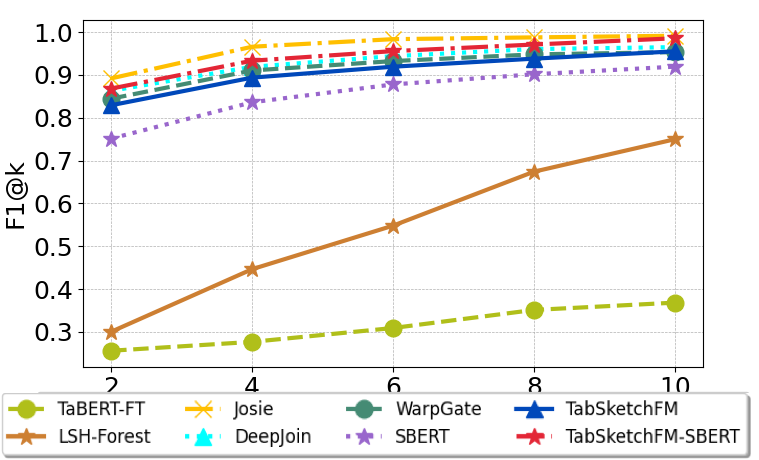}
    \caption{F1 on \textit{wiki join} search for varying \textit{k}.}
    \label{fig:wikijoin_search}
\end{subfigure}
\hfill
\begin{subfigure}{\columnwidth}
    \centering
    \includegraphics[width=0.8\columnwidth]{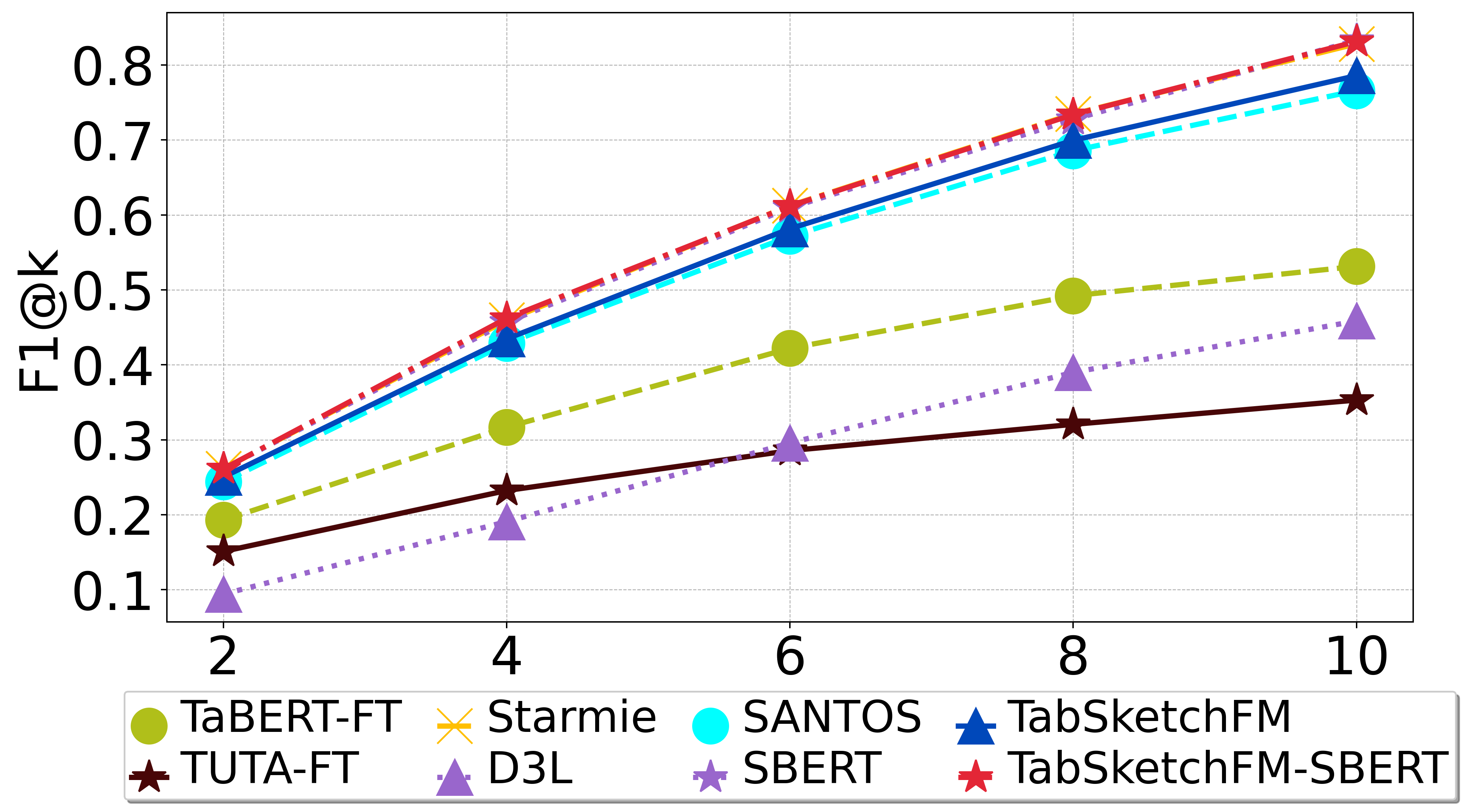}
    \caption{F1 on \textit{Santos} search for varying \textit{k}.}
    \label{fig:santos_search}
\end{subfigure}

\begin{subfigure}{\columnwidth}
    \centering  \includegraphics[width=0.8\columnwidth]{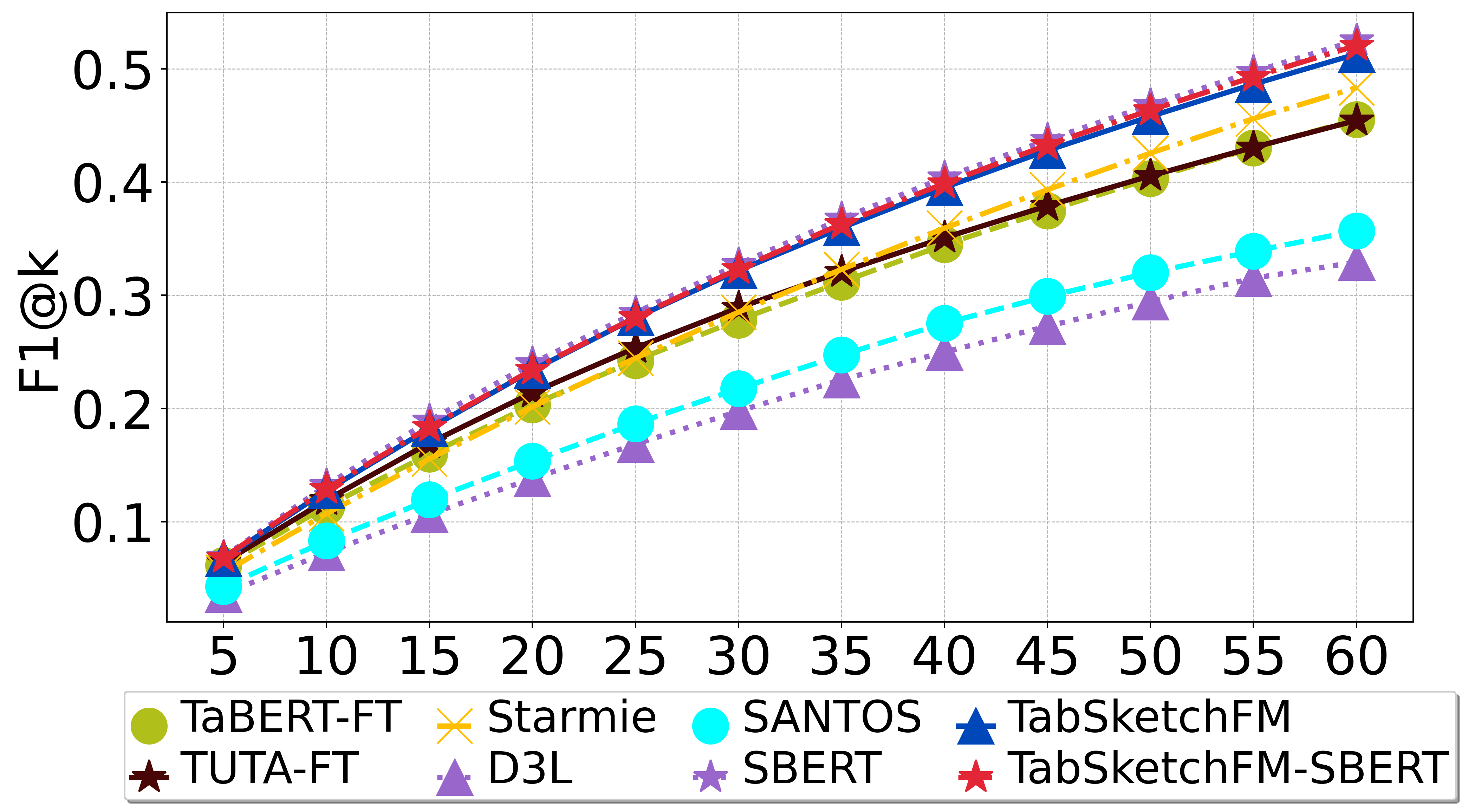}
    \caption{F1 on \textit{TUS} search for varying \textit{k}.}
    \label{fig:tus_search}
\end{subfigure}
\hfill
\begin{subfigure}{\columnwidth}
    \centering  \includegraphics[width=0.8\columnwidth]{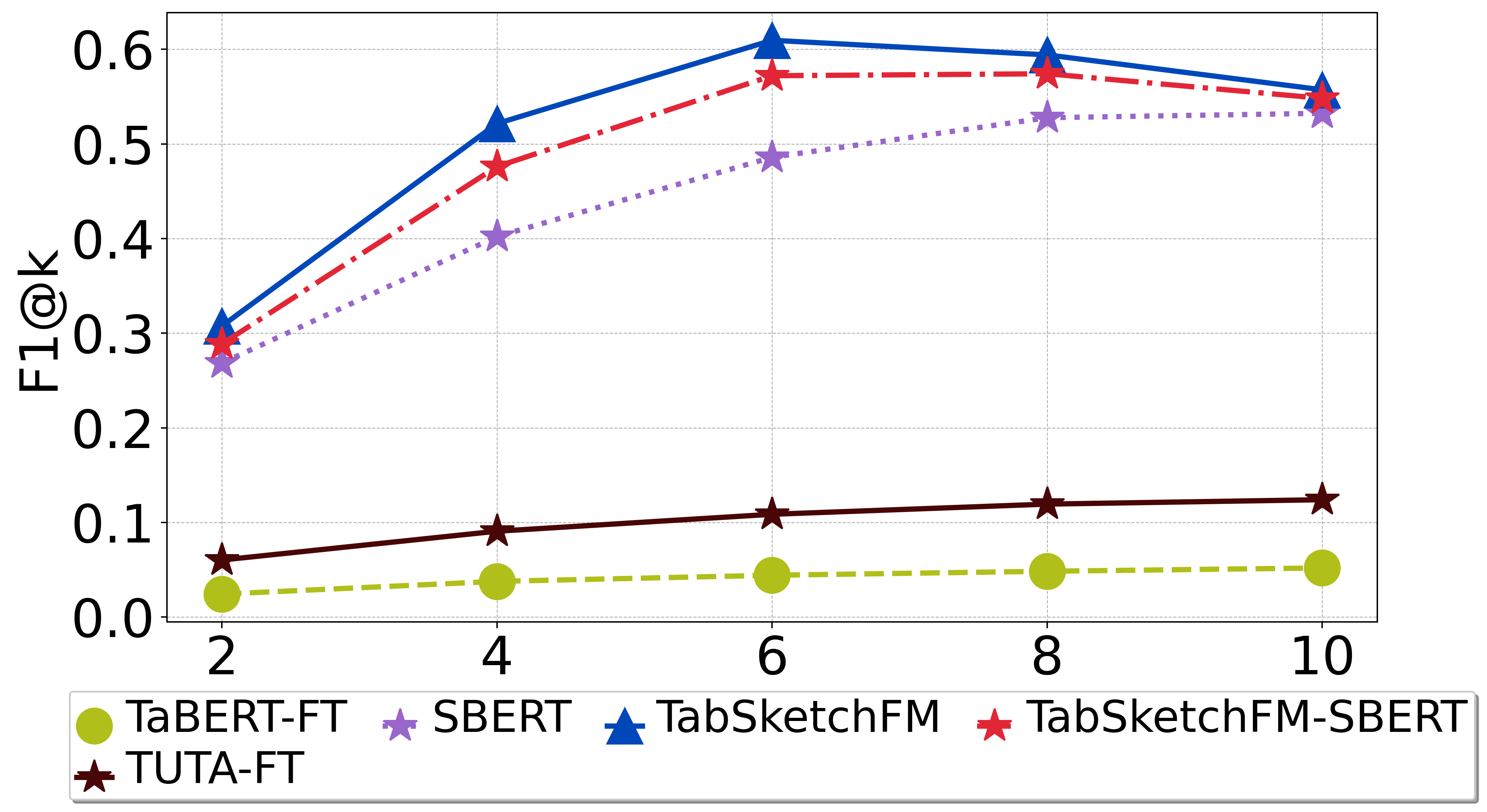}
    \caption{F1 on \textit{Eurostat} subset search for varying \textit{k}.}
    \label{fig:eurostat_search}
\end{subfigure}
\label{}
\caption{F1 plot for search on various join, union and subset datasets.}

\end{figure*}

\begin{figure*}
\begin{minipage}[b]{\columnwidth}
    \captionof{table}{F1, Precision \& Recall for Wiki-join search. }
    \centering
%     \begin{tabular}{l|c|c|c}
% \textbf{Baseline}   & \textbf{Mean F1} & \textbf{P@10} & \textbf{R@10} \\ \hline
%        TaBERT-FT  & 5.88 & 0.43 & 0.04\\
%        LSH-Forest & 10.48 & 0.8 & 0.08\\
%        Josie & 19.56 & \textbf{0.99} & 0.12 \\
%        DeepJoin & 18.88 & 0.96 &  0.11\\
%        WarpGate & 18.58 & 0.95 & 0.11\\
%        SBERT & 83.67 & 0.95 & 0.89 \\ \hline
%        TabSketchFM & \underline{89.09} & 0.97 &  \underline{0.94}\\
%        TabSketchFM-SBERT & \textbf{92.81} & \underline{0.98} & \textbf{0.99} \\
%     \end{tabular}
{ 
    \begin{tabular}{l|c|c|c}
\textbf{Baseline}   & \textbf{Mean F1} & \textbf{P@10} & \textbf{R@10} \\ \hline
       TaBERT-FT  & 30.16 & 0.43 &  0.32 \\
       LSH-Forest & 50.84 & 0.8 &  0.7 \\
       Josie & \textbf{94.86} & \textbf{0.99} & \textbf{1.00} \\
       DeepJoin & 91.59 & 0.96 &  0.97 \\
       WarpGate & 90.34 & 0.95 &  0.95 \\
       SBERT & 83.67 & 0.96 &  0.89 \\\hline
       TabSketchFM & 89.09 & 0.97 &  0.94 \\
       TabSketchFM-SBERT & \underline{92.81} & \underline{0.98} & \underline{0.99}\\
    \end{tabular}
}
    \label{tab:wikijoin_search}
\end{minipage}
\hfill 
 \begin{minipage}[b]{\columnwidth}
    \centering
% \begin{table}
    \captionof{table}{F1, Precision \& Recall for SANTOS union search.}
    \centering
    \begin{tabular}{l|c|c|c}
     \textbf{Baseline}   & \textbf{Mean F1} & \textbf{P@10} & \textbf{R@10}\\ \hline
       TaBERT-FT  & 36.64 &0.63&0.46\\
       TUTA-FT  & 25.34 &0.43&0.3\\
       Starmie & \textbf{54.08} &\textbf{0.97}&\underline{0.72} \\
       D3L & 26.44 &0.54& 0.4 \\ 
       SANTOS & 50.36 & 0.89 & 0.67\\ 
       SBERT & \underline{53.86} & \textbf{0.97}&\textbf{0.73} \\ \hline
       TabSketchFM & 51.38 &\underline{0.92}&0.69\\
       TabSketchFM-SBERT & \textbf{54.09} &\textbf{0.97}&\textbf{0.73}
    \end{tabular}
    \label{tab:santos_search}
% \end{table}
\end{minipage}
\vspace{2em}

\begin{minipage}[b]{\columnwidth}
% \begin{table}
    \centering
     \captionof{table}{F1, Precision \& Recall for TUS union search. $^{\star}$SANTOS results are reported only on 125 queries.}    
    \begin{tabular}{l|c|c|c}
     \textbf{Baseline}   & \textbf{Mean F1} & \textbf{P@60} & \textbf{R@60}\\ \hline
TaBERT-FT & 28.05 & 0.9 &  0.32 \\
TUTA-FT & 28.68 & 0.89 &  0.33 \\
Starmie & 28.79 & 0.9 &  0.33 \\
D3L & 20.77 & 0.6 &  0.23 \\
SANTOS$^{\star}$ & 24.27 & 0.81 &  0.27 \\
SBERT & \textbf{32.73} & \textbf{0.99} &  \textbf{0.38} \\ \hline
TabSketchFM & 32.0 & \underline{0.97} &  \underline{0.37} \\
TabSketch-SBERT & \underline{32.3} & \textbf{0.99} &  \textbf{0.38} \\
       \end{tabular}

    \label{tab:tus_search}
% \end{table}
\end{minipage}
\iffalse
\begin{minipage}[b]{\columnwidth}
% \begin{table}
    \centering
     \captionof{table}{F1, Precision \& Recall for TUS union search. [Reported]}
    \begin{tabular}{l|c|c|c}
     \textbf{Baseline}   & \textbf{Mean F1} & \textbf{P@60} & \textbf{R@60}\\ \hline
       TaBERT-FT  & 26.66 & 0.90&0.30 \\
       TUTA-FT  & 27.27 &0.89&0.31\\
       Starmie & 27.48&0.96& 0.32\\
       D3L & 18.98 &0.75&0.21\\ 
       SANTOS & 20.83 &0.81&0.23\\
       SBERT & \textbf{31.13}&\textbf{0.99} &\textbf{0.36}\\ \hline
       TabSketchFM & 30.43 &\underline{0.97}&\underline{0.35} \\
       TabSketchFM-SBERT & \underline{30.72} &\textbf{0.99}&\underline{0.35}
    \end{tabular}
        \label{tab:tus_search}
    \end{minipage}

\begin{minipage}[b]{\columnwidth}
% \begin{table}
    \centering
     \captionof{table}{F1, Precision \& Recall for TUS union search. [With 125 queries]}    
    \begin{tabular}{l|c|c|c}
     \textbf{Baseline}   & \textbf{Mean F1} & \textbf{P@60} & \textbf{R@60}\\ \hline
TABERT & 27.92 & \underline{0.98} &  \underline{0.33} \\
TUTA &27.54 & 0.96 &  \underline{0.33} \\
STARMIE &27.49 & 0.96 &  0.32 \\
D3L & 22.11 & 0.75 &  0.25 \\
SANTOS &  24.27 & 0.81 &  0.27 \\
SBERT & 28.28 & \textbf{1.0} &  \textbf{0.34} \\ \hline
TabSketchFM & \underline{28.3} & \textbf{1.0} &  \textbf{0.34} \\
TabSketch-SBERT &\textbf{28.31} & \textbf{1.0} &  \textbf{0.34} \\
       \end{tabular}

    \label{tab:tus_search}
% \end{table}
\end{minipage}
\fi
\hfill
\begin{minipage}[b]{\columnwidth}
    \captionof{table}{F1, Precision \& Recall for Eurostat subset search.}
    \centering
    \begin{tabular}{l|c|c|c}
\textbf{Baseline}   & \textbf{Mean F1} & \textbf{P@10} & \textbf{R@10}\\ \hline
       TABERT-FT  & 4.03 &0.05& 0.05\\
       TUTA-FT & 9.82 &0.13&0.12\\
       SBERT & 43.12 &0.56&0.51\\ \hline
       TabSketch & \textbf{49.96} &\textbf{0.59}& \textbf{0.53}\\
       TabSketch-SBERT & \underline{47.54} &\underline{0.58}& \underline{0.52} \\
    \end{tabular}

    \label{tab:eurostat_search}
\end{minipage}
\end{figure*}

\subsubsection{Join Search}
\label{section:join_search}
Here, the task is to find all the tables from the data lake that can be joined on one or more columns with the given query table. 
Note that prior join works~\cite{2019_zhu_josie, 2022_esmailoghli_mate, 2021_dong_pexeso_table_discovery} evaluate the efficiency of searching for data lake columns having value overlap with the query column(s), and assume that the columns with overlapping values are always joinable. However, we are interested in whether it makes sense to join tables.
For example, two columns: (i) explaining people's Ages with values such as (50, 24, 66, 78, 92, 95) and (ii) explaining students' marks out of 100 (rounded to integer) with values such as (78, 92, 95, 100, 20) can have overlapping values but it is not sensible to join them because they have different semantics.
Because \ourmethod{} is built to generate contextual embeddings of columns that include value overlap, column headings, its context with other columns, and so on, it should be possible to get rid of such useless joins.
%in contrast to prior work on joins, which has focused exclusively on efficient ways to determine value overlap over a large data lake. 
%The scalability works generally consider the set of columns in the data lake rather than actual tables~\cite{2019_zhu_josie}. 
%As we use the context of nearby columns in our work, we need a benchmark with the tables in the data lake rather than independent columns.

\begin{figure*}[t]
    \centering
    \includegraphics[width=0.8\textwidth]{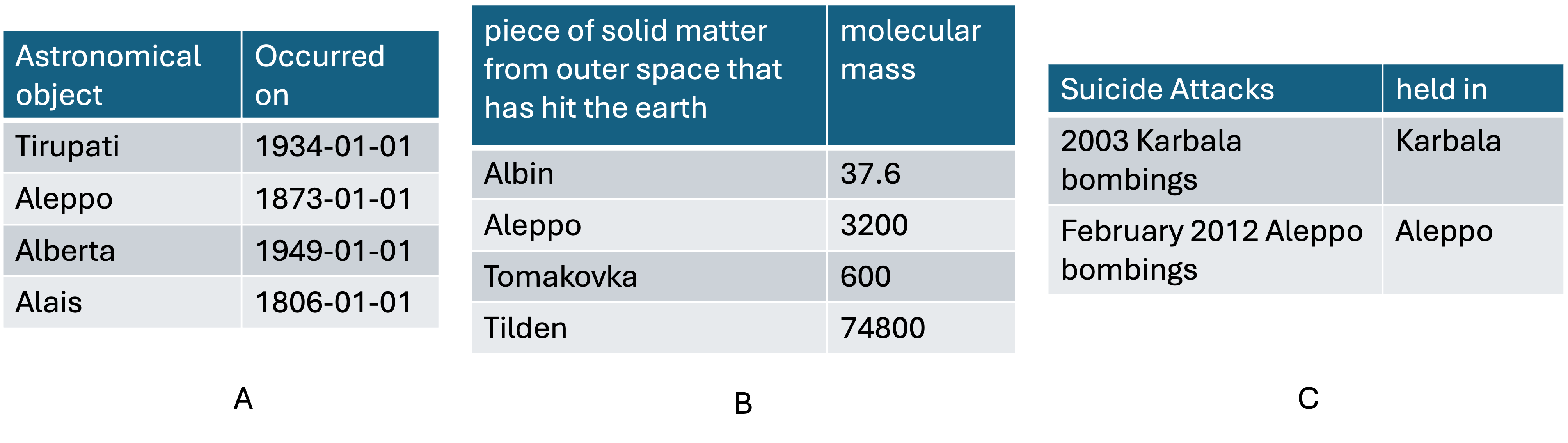}
    \caption{Sample subsets of four tables from our Wikidata-based table union and join benchmark (top), along with ground truth labels and mappings (bottom)}
    \label{fig:wiki_examples}
\end{figure*}

To our knowledge, there are no such benchmarks with annotated gold standards that consider if it is sensible to join\footnote{A new benchmark very recently got published on this topic but that was after we had built the benchmark \cite{vldblakebench}}. 
So, we construct a search benchmark named \textbf{Wiki Join} from Wikidata (by following the cell-entity mapping 
approach used in the curation of the Wiki Jaccard dataset \cite{lakebench}, see Table~\ref{tab:benchmark_cardinality} for dataset characteristics). 
%Such mappings help us to identify if the join is sensible. 
% Specifically, we follow two step process to generate this dataset. First, 
We first generated the tables and then used the ground truth cell-entity mapping to create a list of all the pairs of columns with the same entity annotations. For example, consider Figure~\ref{fig:wiki_examples}. All entities in first column of table A and first column of table B are mapped to \texttt{Meteorite} where as entities in table C are mapped to \texttt{Locations}. So even when there is an overlap in cell values of all tables (\texttt{Aleppo} appears in all the tables), only table A and B are considered sensible to join. Once the list of sensible joinable column pairs is obtained, we assign each pair an overlap
score, which is the Jaccard score of the entity annotation sets. The pairs with overlap score greater than $0.5$ were considered as joinable in our dataset for a total of 5,512 queries. 
% \hk{Do we plan to make it public?}
% As the column pairs are annotated to the same entity it is sensible to join on these columns. 
% We assign each pair an associated overlap score, which is the Jaccard score of cell entity .  For example, in Figure~\ref{fig:wiki_examples}, {\tt col0} in table {\tt T1GB2LNKTOXP.csv} is joinable with {\tt col0} in table {\tt SW4T7EFEI2DK.csv} with a score of $0.5$ (3 overlapping entities out of the 6 entities in both columns).
% For each column in the list of pairs, we create a ranked list of joinable columns 
%of joinable columns 
% in descending order of their scores. 
This dataset contains contains $46,521$ tables and  $5,512$ query columns.
%The overlap score is the jaccard score of the entity annotation sets.
%As with LakeBench, this benchmark will be made available.

%While most literature in the table join search focuses on improving the time efficiency to process the queries for large-scale data lakes; our focus is on evaluating the utility of the finetuned cross-encoder for improving retrieval in terms of precision. 
%Similar to union search, 
We compare the join search performance of \ourmethod{} with two traditional join search methods---\textbf{LSHForest} and \textbf{Josie} \cite{2019_zhu_josie}---and three neural methods--\textbf{WarpGate}~\cite{2023_cong_warpgate}, \textbf{DeepJoin}~\cite{2023_dong_deepjoin}, and \textbf{TaBERT}.  
%
% A recent embedding-based join search system, WarpGate~\cite{2023_cong_warpgate}, is not available publicly. So, we implemented a similar Glove-embedding-based baseline~\cite{2014_pennington_glove} (\textbf{EmbedJoin}) that embeds columns into an embedding space and performs a search over HNSW index~\cite{2020_malkov_hnsw}.  Other openly available systems, such as MATE \cite{10.14778/3529337.3529353} were not easy to install due to software versioning requirements that we could not resolve.  
WarpGate is a recent embedding based method that performs joinable column search using SimHash LSH in the embedding space. It uses FastText embedding pretrained on WebTables to generate embeddings of column values.
DeepJoin is a similar embedding base approach that finetunes a pretrained language model and uses HNSW~\cite{2020_malkov_hnsw} for indexing and efficient search. They experiment with FastText, BERT, DistilBERT and MPNet. In our experiments we use the FastText embedding as it was found to be the best performing embedding when used without finetuning. In addition to column values, DeepJoin also uses column names, table names and column statistics (max, min and average character length) while embedding; much like our sketches. For TaBERT baseline, we use the column embeddings obtained from the TaBERT model which was finetuned on the Wiki-Containment dataset from LakeBench in Section~\ref{exp:finetuning} ({\bf TaBERT-FT}). We could not include TUTA as it does not provide column embeddings. 

% In addition to these join benchmarks, we also 
Recognizing the power of pretrained models trained on massive corpora of text, we added an off-the-shelf sentence encoder \textbf{SBERT} (\texttt{all-MiniLM-L12-v2}) as a powerful baseline. We include a very simple approach of concatenating the top $100$ unique values in a column into a single sentence and encoding it to produced a column embedding. As seen in figure~\ref{fig:wikijoin_search} this simple baseline was quite effective compared to many traditional hand crafted systems for data discovery, as well as other neural systems. 
For finding joins (and subsets), value overlap is a requirement, and data sketches do a very efficient job of finding overlap in large sets, whereas encoding many thousands of values in a sentence is inefficient and likely ineffective.  To examine if there is any gain in combining the embeddings of columns derived from TabSketchFM models along with value embeddings for each column using sentence encodings, we also concatenated the two embeddings after normalizing them so the means and variances of the two vectors were in the same scale (\textbf{TabSketchFM-SBERT}).

% \hk{TBD HARSHA please add on Deepjoin.}  
% { \color{red}
Figure~\ref{fig:wikijoin_search} shows an advantage of \ourmethod{} over other join search baselines in Wiki Join Search benchmark. Table~\ref{tab:wikijoin_search} summarizes these results considering Mean F1, Precision @ 10, and Recall @10. 
In this and other benchmarks, TabSketchFM gives comparable performance with the exact match baseline Josie, and is among the top-2 best performing methods throughout the experiments. Specifically, the variation of TabSketchFM that adds SBERT column embeddings (TabSketchFM-BERT) improves performance of join search by around 3 \% in terms of Mean F1.   
% }

%Note that we want to evaluate effectiveness over efficiency in this section, and the recent systems yield the same output but more efficiently. \textbf{LSHForest} is 
%\footnote{\url{https://github.com/ekzhu/datasketch}}, 
%an approximate Jaccard-based join search technique that retrieves top-k data lake tables having joinable columns with the query column~\cite{BawaCG05}.
%\textbf{LSHForest} approximates Jaccard score using MinHash and LSH. 
%We use the implementation of LSHForest in the DataSketch package~\cite{eric_zhu_2017_290602}. 

\begin{figure*}[!ht]
\begin{center}
    $$
    \begin{array}{rcl}
        \textsc{KnnSearch}\left(c, k\right) & \equiv & \left(\left<c_i, d_i\right>\dots\right) \; \text{$\left(k \times 3\right)$ nearest columns $c_i$ by distance $d_i$}\\
        \textsc{ColumnNearTables}\left(c, k\right) & \equiv &  \bigcup\limits_{t_i \in \left\{ \textsc{Table}(c_i) \left| c_i \in \textsc{KnnSearch}(c, k) \right. \right\}}
          \left< t_i, \min\limits_{\left<c_i, d_i\right> \in \textsc{KnnSearch}\left(c, k\right) \wedge \textsc{Table}(c_i) = t_i} d_i \right>\\
        \textsc{NearTables}(t) & \equiv & \left\{ \left<t_i, \left\{ \left<c_i, d_i\right> \right\} \right>  \left| \left< \left<c_i, d_i\right> \in \bigexists_{c \in \textsc{ColumnNearestTables}(t, k)} \textsc{ColumnNearestTables}(c, k) \right. \right> \right\} \\
        \textsc{Rank}_1\left(\left<t_i, \left\{ \left<c_i, d_i\right> \right\} \right> \right)  & \equiv & \left| \left\{ c_i \dots \right\} \right| \\
        \textsc{Rank}_2\left(\left<t_i, \left\{ \left<c_i, d_i\right> \right\} \right> \right)  & \equiv & \sum d_i\\
    \end{array}$$
\end{center}
\vspace{-2em}
\caption{Ranking of nearest tables}
\label{nearestTables}
\end{figure*}

\subsubsection{Union Search}

%In \emph{union search}, the task is to find all the tables from a given collection of tables in a data lake that are unionable with the query table~\cite{Nargesian2018_Table_union}. 
%
%As in previous works~\cite{2023_fan_starmie, 2023_hu_autotus}, 
We use two publicly available benchmark datasets: $\mathbf{TUS_{Small}}$ and $\mathbf{SANTOS_{Small}}$~\cite{Nargesian2018_Table_union, Khatiwada2022_SANTOS} because the larger versions of the dataset are not labeled.
%For stage one of the process, 
As a baseline, we compare against state-of-the-art table union search techniques: $\mathbf{D^3L}$~\cite{2020_bogatu_d3l}, \textbf{SANTOS}~\cite{Khatiwada2022_SANTOS}, and \textbf{Starmie}~\cite{2023_fan_starmie} as well as the \textbf{SBERT} baseline explained earlier. 
%, to obtain top-100 candidates for each query. 
%
% We report results on ${D^3L}$%~\cite{2020_bogatu_d3l} 
% % \footnote{\url{https://github.com/alex-bogatu/d3l}}
% , 
% {SANTOS}%~\cite{Khatiwada2022_SANTOS}
% % \footnote{\url{https://github.com/northeastern-datalab/santos}}
% , and 
% \textbf{Starmie}%~\cite{2023_fan_starmie}
% % \footnote{\url{https://github.com/megagonlabs/starmie}}
% ~using their publicly available code. 
We could not compare against $AutoTUS$~\cite{2023_hu_autotus}, a recent union search system, because its code was not public at the time of writing this paper.

To determine unionability, table embeddings from \ourmethod{} did not yield strong results; perhaps because the union benchmarks had been created by creating subsets of the columns in a table or rows.
Following the state-of-the-art unionability approach Starmie\cite{2023_fan_starmie}, we used column embeddings of \ourmethod{} to determine if two tables are unionable. But we used a simpler technique than the bipartite graph matching algorithm introduced by Starmie. Our approach is defined in Figure~\ref{nearestTables}, and works as follows:
\begin{enumerate}
    \item For each column $c$ in a given table $t$, $\textsc{KnnSearch}$ returns the nearest $k * 3$ {\em columns} in the dataset.  Note that we try to get a lot more columns than $k$ simply to increase the number of candidate columns we consider at this step because multiple columns from a single table might match a given column.
    \item For a column $c$, $\textsc{ColumnNearestTables}$ returns each table near $c$ and the distance of its closest column.
    \item For a table $t$, $\textsc{NearestTables}$ gathers the near tables for each column in $t$.
    \item We then choose the best, first using $\textsc{Rank}_1$ to select the tables with the largest number of matching columns, and then, for ones where that is equal, $\textsc{Rank}_2$ chooses the closest, i.e. the smallest sum of column distances.
\end{enumerate}

For each technique, we use default parameters suggested in the respective papers.  For TUTA we had to use table embeddings because that was what the model made available.  For both the TUTA and TaBERT, we use the model that was fine tuned on TUS-SANTOS dataset from LakeBench in Sec.~\ref{exp:finetuning}(denoted by {\bf TUTA-FT} and {\bf TaBERT-FT}).  The results for \ourmethod{} compared to the baselines show that \ourmethod{} matched the best system \cite{2023_fan_starmie} on SANTOS, when we added embeddings from SBERT, see Figure~\ref{fig:santos_search}, and Table~\ref{tab:santos_search} for a summary across methods.  For TUS, following the literaure~\cite{Khatiwada2022_SANTOS, 2023_fan_starmie, 2023_hu_autotus}, we report the results for tables having at least 60 unionable tables as we consider the value of k up to 60. Additionally, as SANTOS required manual annotation of intent, we only run this on a subset of $\mathbf{TUS_{Small}}$ queries used in the original paper~\cite{Khatiwada2022_SANTOS}. The value embeddings from the off-the-shelf SBERT model performed very well, as did \ourmethod{} when enhanced with value embeddings; see Figure~\ref{fig:tus_search} and Table~\ref{tab:tus_search}. This suggests for union, column values alone are sufficient to find good unionable results; table specific embeddings do not seem to add much.  Note that value embeddings out performed all the other systems, pointing to the power of existing pretrained models in embedding tabular data.

\subsubsection{Subset Search}
\begin{figure*}[!ht]
    \centering
    \includegraphics[width=0.6\textwidth]{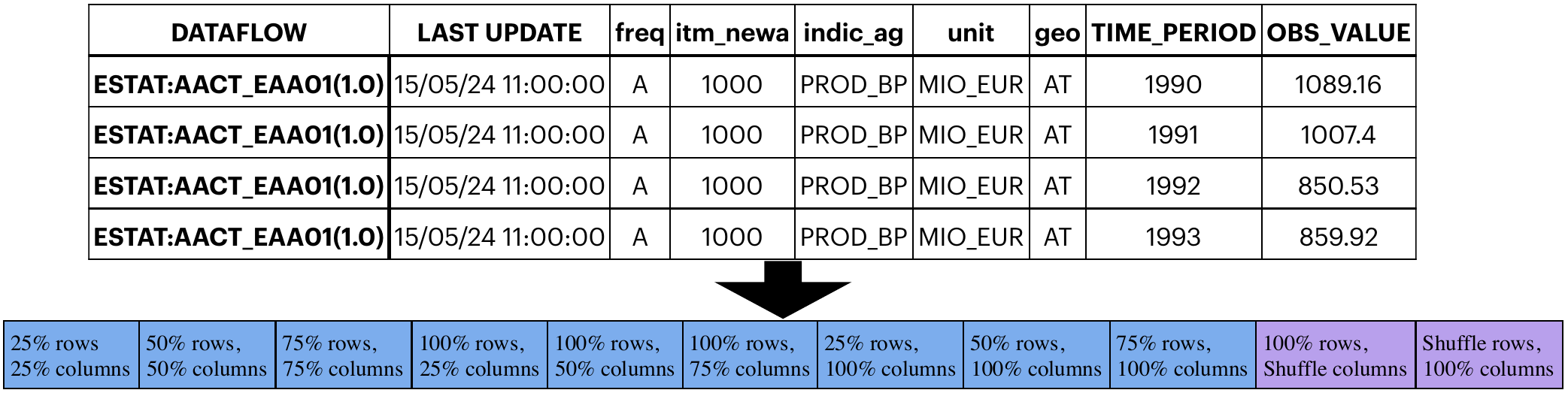}
    \caption{Sample table and generated subsets from Eurostat}
    \label{fig:eurostat}
\end{figure*}

To our knowledge, there are no benchmarks for subset search.  To ensure that the models we built generalize to search, we constructed a dataset for subset search, from Eurostat \footnote{\url{https://data.europa.eu/en}} of 3,072 CSV files comprising the variety of data collected by the EU.  These files serve as query tables for search.  Data characteristics for this benchmark are provided in Table~\ref{tab:benchmark_cardinality}.
For each file, as shown in Figure~\ref{fig:eurostat}, we created the following variants reflecting different types of subsets of the file, sampling 25\%, 50\% or 75\% or all of the rows or columns in the table.  When all rows or columns were kept, we shuffled either the rows or the columns (shown in purple).  These last two variants were added to measure whether the embeddings capture table semantics, that is order invariance for rows and columns of a table. 
 The 11 variants per file made up a search dataset of 38,904 files.

\begin{figure*}[!ht]
\begin{subfigure}{\columnwidth}
\centering
    \includegraphics[width=0.8\columnwidth]{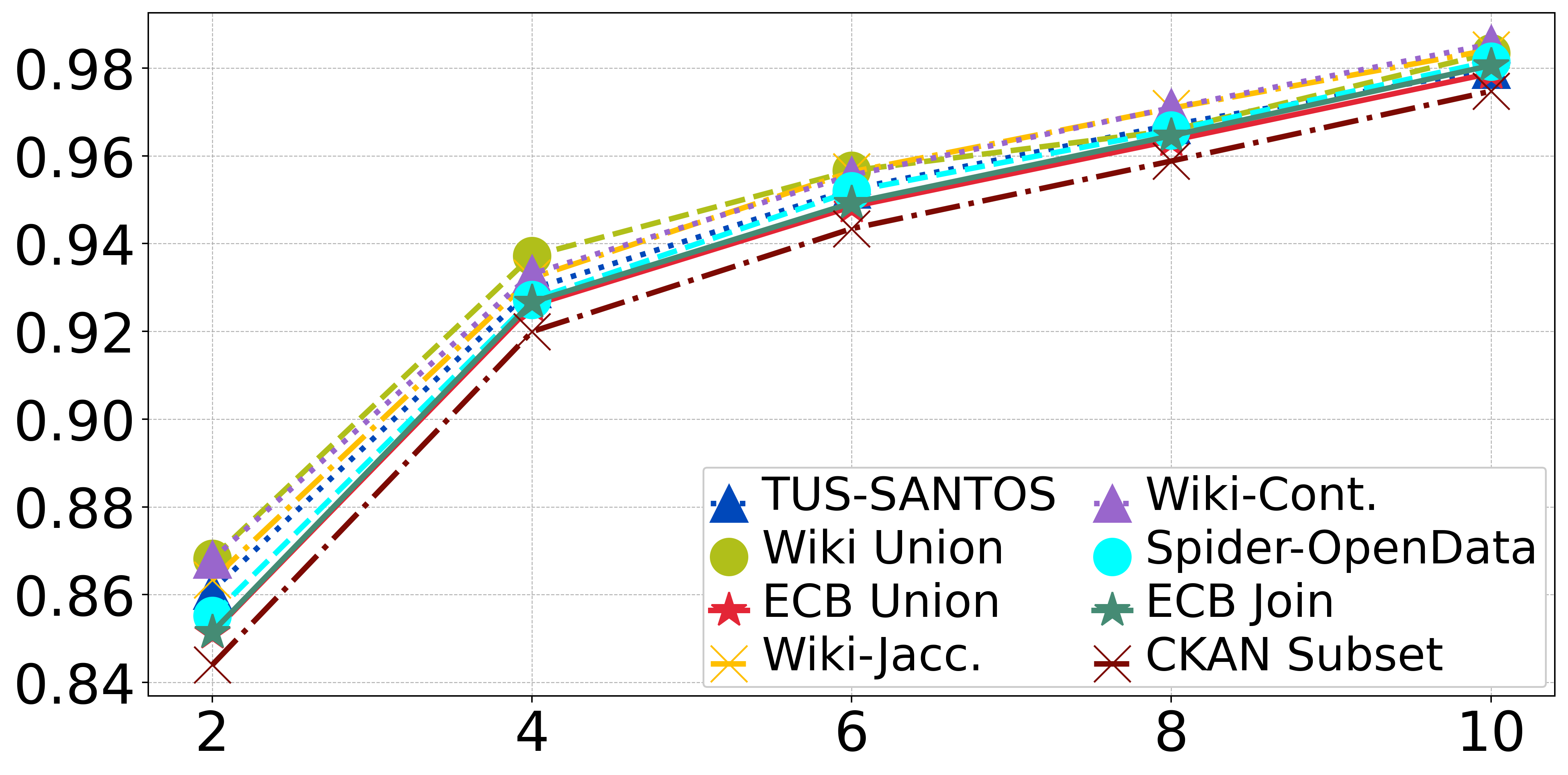}
    \caption{F1 on \textit{wiki join} transfer on search for varying \textit{k}.}
    \label{fig:wikijoin_transfer}
\end{subfigure}
\hfill
\begin{subfigure}{\columnwidth}
\centering
    \includegraphics[width=0.8\columnwidth]{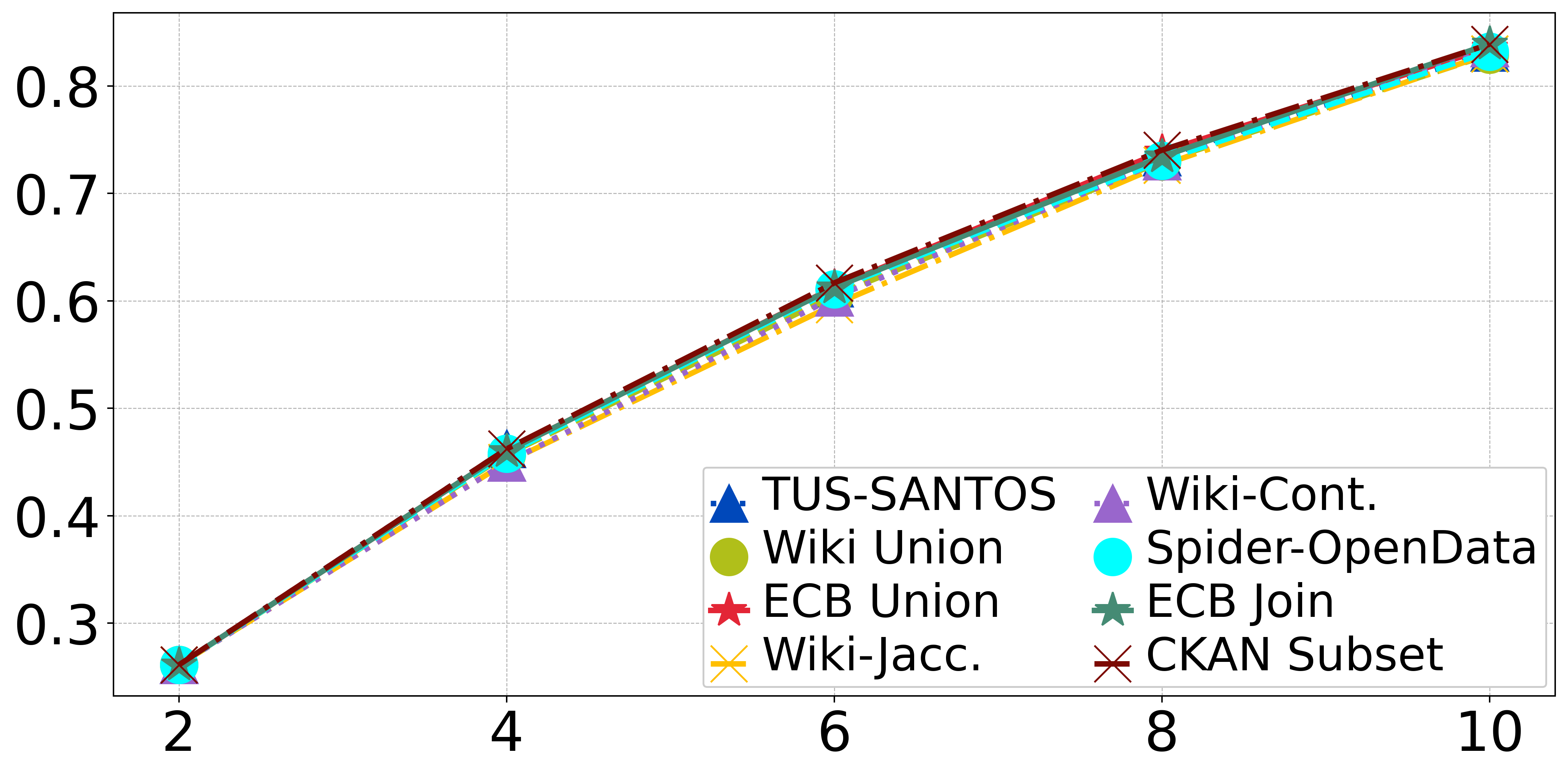}
    \caption{F1 on \textit{Santos} transfer on search for varying \textit{k}.}
    \label{fig:santos_transfer}
\end{subfigure}

\begin{subfigure}{\columnwidth}
\centering
      \includegraphics[width=0.8\columnwidth]{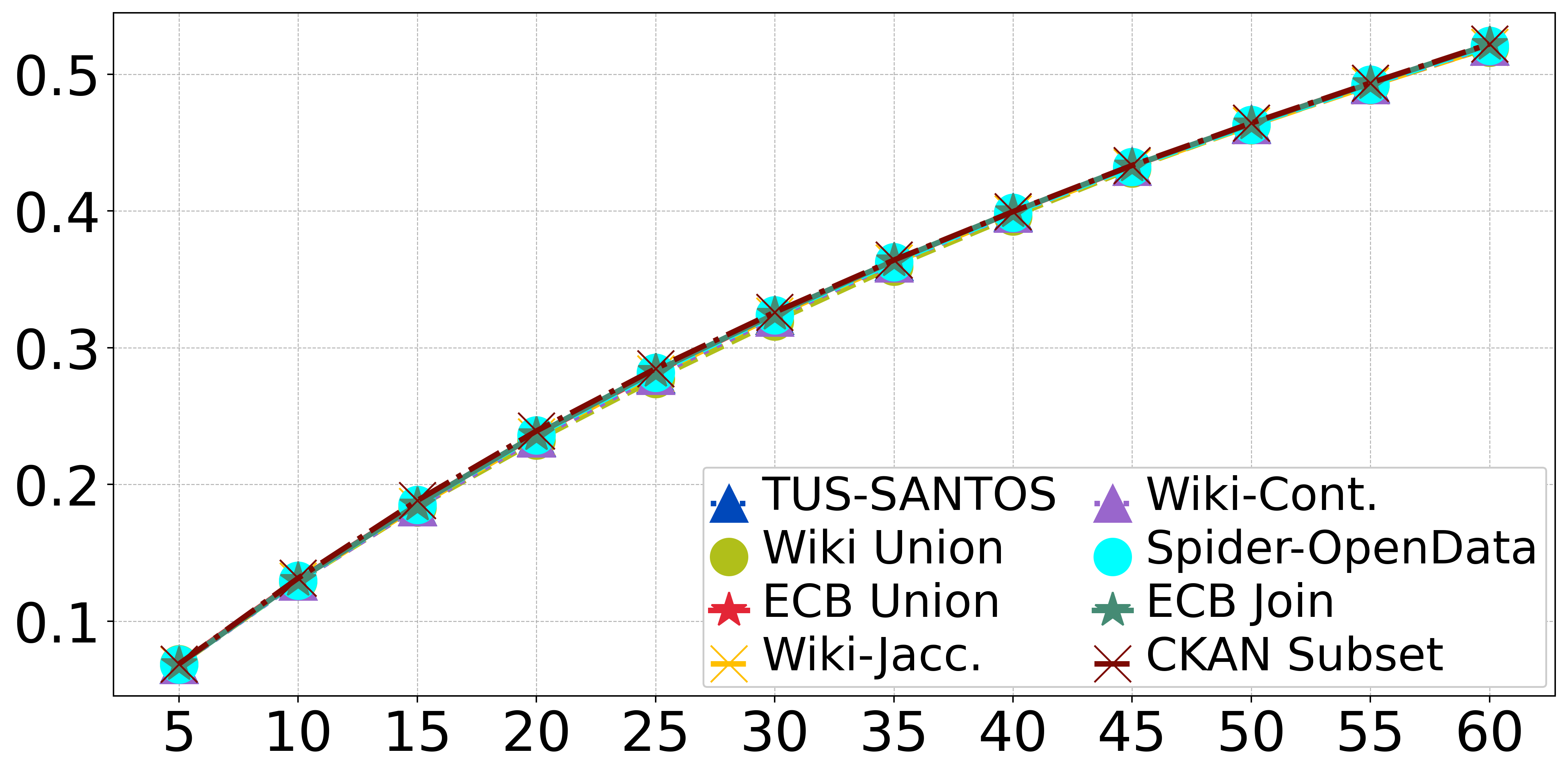}
    \caption{F1 on \textit{TUS} transfer on search for varying \textit{k}.}
    \label{fig:tus_transfer}
\end{subfigure}
\hfill
\begin{subfigure}{\columnwidth}
    \centering  
    \includegraphics[width=0.8\columnwidth]{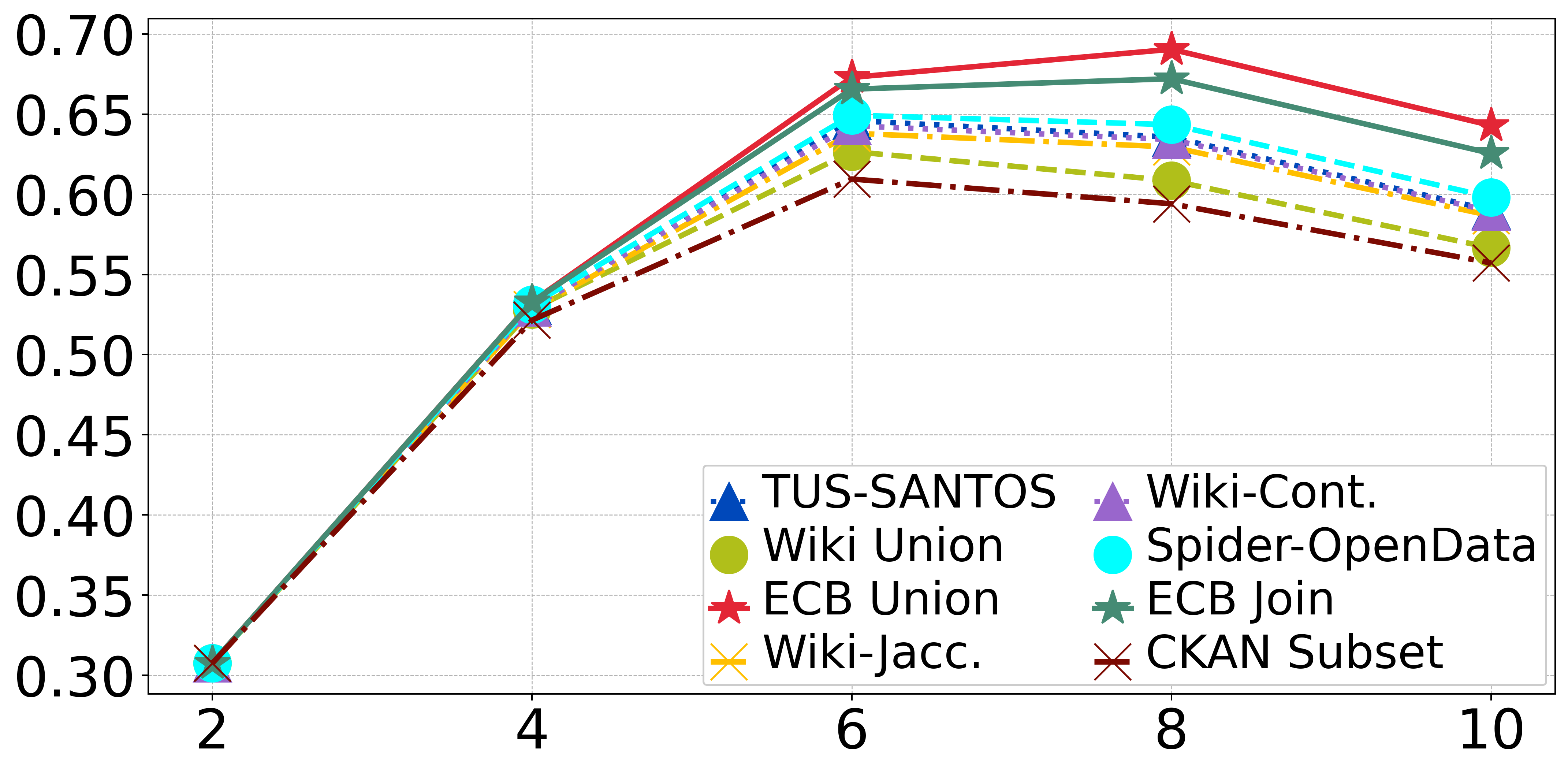}
    \caption{F1 on \textit{Eurostat} transfer on search for varying \textit{k}.}
    \label{fig:eurostat_transfer}
\end{subfigure}
\caption{F1 plots for transfer across different K values. (a) WikiJoin (b) Santos (c) TUS (d) Eurostat}
\label{}
\end{figure*}

We follows the same procedure described in Figure~\ref{nearestTables} to find tables that were subsets of the query tables.  Figure~\ref{fig:eurostat_search} and Table~\ref{tab:eurostat_search} shows the results of the subset search on the fine tuned model trained on ckan-subset.  As we see in the next section on transfer, this model turned out to be the weakest TabSketchFM models for the subset search problem, likely reflecting the quality of the dataset used for training The best performing TabSketchFM model was trained on union, on the European Central Bank data at 0.55.  Still, we include it for consistency.  In addition, the results suggest that for subsets, adding value embeddings of columns from SBERT actually made performance slightly worse.  This highlights the importance of being able to customize embeddings for different data discovery tasks - the ability to add value embeddings to tabular embeddings seems to be useful for such customization.

One additional question we asked is whether the TabSketchFM models exhibited column order invariance, and row order invariance.  Unsurprisingly, changing row order made no difference to TabSketchFM, all 3072 queried tables returned the row shuffled variants in their nearest neighbors set. In contrast, changing row order did influence SBERT embeddings; only 2809/3072 (91\%) row shuffled variants were returned. Column order can matter to TabSketchFM models, since the encoder is a sequence model.  For the column shuffled variants, 3059/3072 appeared in the nearest neighbors set (99.5\%).  The SBERT model of course did not consider the larger column context, and was hence unaffected by the shuffling of columns. 

Overall, from our evaluations of $4$ data discovery datasets across traditional and neural approaches answers~\ref{Q3}, that embeddings from finetuned \ourmethod{} can be used for data discovery tasks. As the embeddings are pretty small vectors the search in embedding space can be made extremely efficient; making it an ideal candidate for large data lakes.

%Together, the answer to our last question (\ref{Q3}) suggests that neural models, like ours, can be used to rerank the search results. 
\subsubsection{Generalization across tasks and domains}

We now investigate~\ref{Q4}. How does our method, fine-tuned on a single task and domain, can adapt and generalize to a new domain.
Figures~\ref{fig:wikijoin_transfer}, \ref{fig:santos_transfer}, and \ref{fig:tus_transfer}, \ref{fig:eurostat_transfer},  show that the fine tuned cross encoder models generalize rather well across tasks as well as domain, which is encouraging particularly because it shows that the models can be trained on a datalake that is quite different from the datalake that it is deployed to.  In enterprise contexts, this is an important requirement, because it means that we can basically train models offline and apply it online.  All models shown in the Figure are models that include the value embeddings from a column for maximal generalization, but using the models without values also produced the same type of generalization.  

%% file: sections/conclusion.tex
\section{Conclusion and Discussion}

We presented \ourmethod, a novel transformer-based tabular representation model that inputs the table in the form of several sketches. We show that cross encoders based on \ourmethod{} outperform existing systems on LakeBench related table identification task. We also show that finetuned \ourmethod{} can be effectively used to search for tables that are unionable, joinable or are subsets of each other. We highlight that embedding values using sentence encoder models can be a useful addition for some discovery tasks. As far as the search efficiency is concerned, we recommend indexing the datalake offline and at query time only compute embeddings for the query table. This computation is similar to other embedding-based search system. 

%We created new benchmarks for the dataset discovery task and evaluated the existing tabular models against them. As is evident from the empirical results, most tabular models, being trained on web tables, did not handle large enterprise tables with limited metadata well. 
% Since dataset discovery is an important problem and there is a lack of datasets to train neural models for this task, we believe that LakeBench would be a useful resource for building better neural models in this space.